%% file: elsarticle-template-1-num.tex
\documentclass[preprint,12pt]{elsarticle}

\usepackage{graphicx}
\usepackage{caption}
\usepackage{subfigure}
\usepackage{amssymb}
\usepackage{diagbox}
\usepackage{float}
\usepackage{lineno}
\usepackage{subfig}
\usepackage{algorithm2e}
\usepackage{amsmath}

\usepackage{color,xcolor}

\usepackage{booktabs}
\usepackage{bm}
\usepackage{makecell}
\usepackage{multirow}
\usepackage{amsmath}
\usepackage{mathrsfs}
\usepackage{amssymb}
\usepackage{natbib}
\usepackage{rotating}
\journal{Structural Health Monitoring}

\begin{document}

\begin{frontmatter}


\title{Analysis for full face mechanical behaviors through spatial deduction model with real-time monitoring data}


% \author[label1,label2]{<author name>}
% \address[label1]{<address>}
% \address[label2]{<address>}
\author[1,3]{Xuyan Tan}
\author[2]{Yuhang Wang}
\author[2]{Bowen Du\corref{cor1}}
\ead{dubowen@buaa.edu.cn}
\author[2]{Junchen Ye}
\author[1,3]{Weizhong Chen}
\author[2]{Leilei Sun}
\author[4]{Liping Li}

\address[1]{State Key Laboratory of Geomechanics and Geotechnical Engineering, Institute of Rock and Soil Mechanics, Chinese Academy of Sciences, Wuhan 430071, China}
\address[2]{SKLSDE Lab, Beihang University, Beijing 100083, China}
\address[3]{University of Chinese Academy of Sciences, Beijing 100049, China}
\address[4]{Geotechnical and Structural Engineering Research Center, Shandong University, Ji'nan 250061,  China}
\cortext[cor1]{Corresponding Author.}

\begin{abstract}
Mechanical analysis for the full face of tunnel structure is crucial to maintain stability, which is a challenge in classical analytical solutions and data analysis. Along this line, this study aims to develop a spatial deduction model to obtain the full-faced mechanical behaviors through integrating mechanical properties into pure data-driven model. 
The spatial tunnel structure is divided into many parts and reconstructed in a form of matrix. Then, the external load applied on structure in the field was considered to study the mechanical behaviors of tunnel. Based on the limited observed monitoring data in matrix and mechanical analysis results, a double-driven model was developed to obtain the full-faced information, in which the data-driven model was the dominant one and the mechanical constraint was the secondary one.
To verify the presented spatial deduction model, cross-test was conducted through assuming partial monitoring data are unknown and regarding them as testing points. The well agreement between deduction results with actual monitoring results means the proposed model is reasonable. Therefore, it was employed to deduct both the current and historical performance of tunnel full face, which is crucial to prevent structural disasters.

\end{abstract}


\end{frontmatter}

% \linenumbers


\input{1-Introduction}
\input{2-Background}
\input{3-Methodology}

\input{4-Expriment}

\input{5-Application}
\input{6-Conclusion}
\input{7-Acknowledgement}



%% References



%% New version of the num-names style
% \bibliographystyle{elsarticle-num}
% \bbliography{sample.bib}

\clearpage
\input{Pics}
\clearpage
\input{Tables}






% \end{thebibliography}

\end{document}

%% file: 1-Introduction.tex
\section{Introduction}
\label{S:1}
In past few years, structural health monitoring has been developed to a common technology for stability maintaining\cite{schroder2013approach,annamdas2017applications}. The valid monitoring data provides essential foundation for mechanical behaviors analysis and further performance prediction\cite{salazar2017data}. In order to collect accurate monitoring data, quite a few scholars focused on the research of new sensors and the optimization of monitoring scheme\cite{li2010combined}. Nevertheless, it is impossible to achieve the full face monitoring of structure. Most of the current monitoring sections and monitoring points are determined on the basis of empirical and semi-empirical methods\cite{ariznavarreta2016measurement,li2020analysis,simeoni2009method}. Especially for the constructions located in the complicated underground conditions, many emergencies may occur in the location where there are no monitoring points\cite{yang2018structural}. Thus, it is of great significance to analyze structural mechanical behaviors utilizing the optimized monitoring scheme and limited monitoring points. Along this line, this study aims to perform a novel spatial deduction model using machine learning algorithm to investigate full face mechanical behaviors of structure.

The analysis of structural mechanical behaviors is essential to maintain the stability of structure, which has been studied by classical theoretical mechanics for a couple of years\cite{tan2020settlement,wang2019evaluation}. Due to the difficulty of  solving, the traditional methods are replaced gradually by numerical simulation over the years, such as finite element method, discrete element, and the extension of them\cite{wang2019the,li2015progressive}. The sensitive and dangerous positions of structure can be reflected by numerical results through applying the actual boundary conditions on developed models, which provides vital reference to design monitoring scheme. Therefore, numerical analysis is one of the extensively used methods to determine monitoring sections and monitoring points. However, it is well known that the variation of boundary conditions has significant influence on structural mechanical performance\cite{cao2018mechanical}. The construction is easy to suffer from disturbance of unpredictable emergencies under the hostile environment, which makes the mechanical analysis calculated according to the original boundary condition invalid. It is extremely urgent to seek an advanced method flexible to the complicated boundary conditions. Recently, machine learning methods have been applied to a wide range of daily life with the rapid development of computer science, including transportation planning, medicine research, financial field, to name but a few\cite{ye2019co-prediction,lai2018modeling,yin2016joint}. 
It has become the current trend to exact features of structure variation from mass number of valid data. Compared to the promotion of above fields, the application of machine learning in civil engineering is relatively limited. This is because the traditional methods to collect data in field mainly rely on manual testing, which induced the scarcity of data information. With the promotion of Structural Health Monitoring System (SHMS) in infrastructures, a growing number of scholars focused on adopting the machine learning algorithms to analyze the monitoring data and solve practical problems in civil engineering\cite{salimi2019application}. 
According to whether the data is labeled or not, the machine learning methods are classified into two categories. Supervised learning algorithms include regression models, support vector machine, k-nearest neighbor algorithm and decision trees\cite{wang2015a,poloczek2014knn}, while unsupervised learning algorithms include principal component analysis and k-means\cite{aydilek2013a,gao2020multi-scale}. These methods have been widely used in the application of data augmentation, such as the identification of abnormal data and imputation of missing data\cite{wang2020real,qin2019matrix,zhu2019temperature}. 
Even more, Non-Negative Matrix Factorization (NMF) is a hot topic in machine learning and it has been successfully applied to various tasks, such as structure from-motion problems, acoustic event classification, and recommender systems\cite{xu2020bayesian,LUDENACHOEZ2019214}. NMF is outstanding from many algorithms due to the characteristics of variable dimensions, which means that an observed matrix can be approximately decomposed into two low rank matrices represented positive basis vectors and coefficients weighting \cite{GLOAGUEN2019229,LIANG2020106056}. Inspired by the special properties, the limited time series of monitoring data could be used to deduce a multi-dimensional matrix representing the mechanical variation of more spatial positions. Thus, according to a data-centric viewpoint, a novel spatial-temporal model is presented to analyze the full face mechanical behaviors of structure. As a case study, the presented model is employed in an underwater shield tunnel for experiment. 

This study is structured as followed. Firstly, the background of the case study is introduced. Then, the framework of spatial-temporal model is developed and formalized in the study case. Based on the limited monitoring data, the deduction experiment is carried out under different boundary conditions. Also, the experiment results were discussed and verified by cross-test method. Lastly, the presented model is used to deduce the current stress distribution and historical performance of structure as a promising application. 

%ifferent to the existing researches, the contributions of this study are summarized as follows:
%(1) A spatial deduction model is presented in this study to analyze the full face mechanical behaviors of tunnel structure, including the positions with no sensor monitoring.
%(1) The spatial tunnel structure is partitioned into many elements and formalized in the form of matrix. Every element of tunnel has its corresponding unite in matrix. 

%(2) The proposed model employed analysis of stress boundary conditions into classical pure data-driven model, which is an integration of mechanical analysis with machine learning.
%(2) A spatial deduction model is presented via integrating structural mechanical properties into classical pure data-driven model, which is verified by multiple crossing testing experiment conduced on the basis of real-time monitoring data. 

%(3) The crossing experiments are conducted to verify the presented model. Also, the current stress distribution and structural historical performance are deduced as a promising application.
%(3) The presented model is used to deduct the full face mechanical behaviours of tunnel, including both the current stress distribution and structural historical performance. 

%% file: 2-Background.tex
\section{Background for an underwater shield tunnel}
The Dinghuaimen tunnel, formerly named Nanjing Yangtze River tunnel is the longest shield tunnel constructed under the Yangtze River. The project is located in Nanjing, Jiangsu, China. The landform in the site is mainly flood plain, which is wide, flat and at the altitude of 6-10 meters. Also, the main geological layers encountered by Tunnel Boring  Machine (TBM) are mucky silty clay, silty clay with silty sand, silty-fine sand, medium-coarse sand, gravelly sand, siltstone and pebble. This tunnel consists of two lines, and it is designed as the form of two-tubing dual 8-lane expressways. The maximal length of Dinghuaimen tunnel is 7014m, including the shield-crossing portion with 3537m. 
The lining ring consists of ten segments, i.e., one key segment, two adjacent segments, and seven standard segments. The external diagram is 14.5m, and the width and thickness of segment is 2m and 0.6m, respectively. 

Dinghuaimen tunnel was put into operation in January 1st, 2016. In order to analyze the mechanical behaviors of structure and prevent abnormal conditions, an automatic Structural Health Monitoring System (SHMS) was employed in  this construction and began to service in June 2016, which has been introduced in detail in our previous researches\cite{tan2019a}. Totally ten monitoring sections were determined to install fiber optical sensors according to the structural characteristics and geological conditions. For each monitoring section, twenty stress sensors were pre-buried in segments to have a real-time monitoring of stress variation. The layout of monitoring sections and sensors are displayed in Figure \ref{fig:layout}, where the monitoring sections are numbered from $S1$ to $S10$ .

% \begin{figure}[H]
%     \centering
%     \includegraphics[width=12cm]{pics/layout.png}
%     \caption{Monitoring scheme and system networking in the study site.}
%     \label{fig:layout}
% \end{figure}

%% file: 3-Methodology.tex
\section{Methodology of spatial deduction for tunnel structure}
\label{S:3}
This section focused on the development of spatial deduction model. Specifically, the components of presented models are formalized according to actual conditions in the field.

\subsection{Flowchart of spatial deduction}

Spatial deduction aims to use the few observed data in some positions to deduce the data distributed on the whole space. For tunnel engineering, the full-faced mechanical behavior can be deduced by collecting monitoring data from sensors installed on the section. The flowchart of the proposed spatial deduction algorithm is shown in Figure \ref{fig:flowchart}. The mechanical behaviors where the sensors have been deployed were recorded. The tunnel model is simply reconstructed by a matrix $X$ whose rows represent different radius of the tunnel face and columns represent different layers of the tunnel around the circle. So the location of  sensor can be represented by the subscript of this matrix and the data collected from these sensors can be put into the corresponding subscript to form a sparse matrix where non-empty units correspond to the data collected from the deployed sensors and empty units are what need to be deduced. Due to the unique structure of each tunnel, it is necessary to capture the properties of tunnel structure through boundary conditions analysis. Under the constraint of low rank, the sparse matrix $X$ could be decomposed by mechanical properties. The two factored matrices, $U$ and $V$ could reconstruct a matrix with non-empty units.

% \begin{figure}[h]
% \centering\includegraphics[width=0.7\linewidth]{pics/flowchart.png}
% \caption{Flowchart of the proposed spatial deduction model.}
% \label{fig:flowchart}
% \end{figure}

\subsection{Reconstruction of tunnel model}

Considering tunnel structure is a coherent force body, both the data recorded by sensors and the derived data should be stored in a processable format. Therefore, it is necessary to reconstruct the formal tunnel model. 

The process for reconstructing the tunnel model is shown in Figure 3. In Figure 3(a), tunnel section is firstly divided into $M$ layers. Each layer is divided into $N$ parts along the circumference, which leads to total $M*N$ parts. The parts colored in red represent sensors positions, through which the mechanical variation of these parts are monitored directly. In Figure \ref{fig:reconstruct}, the circle tunnel model is transformed into a form of matrix by flattening out the circular tunnel model clockwise from the arch crown. 
By splitting and transforming, a new matrix $X\in\bm{R^{M \times N}}$ is constructed to represent the formal tunnel model. 
% In matrix $X\in\bm{R^{M \times N}}$, the number of rows is equal to the number of layers, the number of columns is equal to the number of parts in one layer. 
Specifically, the values of non-empty units represent the data collected from the corresponding deployed sensors, and empty units indicate the non-sensor locations where the monitoring values need to be deduced. The problem of full face mechanical behaviors analysis could be defined as finding a mapping function $f$ to fill the empty units with the information in non-empty units:
\begin{equation}
    \hat{X}=f(X),
\end{equation}
where $\hat{X}$ is the deduced matrix.

% \begin{figure}[H] 
%     \begin{minipage}[t]{0.5\linewidth}
%     \centering 
%     \includegraphics[height=4.0cm,width=4.0cm]{pics/3-2.png} {\label{fig:side:a} }
%     \caption*{(a)} 
%     \end{minipage} 
%     \begin{minipage}[t]{0.5\linewidth}
%     \centering 
%     \includegraphics[height=2.8cm,width=3.8cm]{pics/3-4-1.png}
%     \caption*{(b)} 
%     \label{fig:side:b} 
%     \end{minipage} 
%     \caption{The process of reconstructing the tunnel model: (a) division model of formal tunnel, and (b) transform the circle model into a form of matrix.}
% \label{fig:reconstruct}
% \end{figure} 

\subsection{Matrix factorization with tunnel mechanical analysis}

% This section described the method based on Non-negative Matrix Factorization (NMF) to fill up the constructed sparse matrix. Given the matrix $X \in \bm{R^{M \times N}}$, NMF approximates it with two low-rank matrix $U$ and $V$\cite{TENG2019205,LUDENACHOEZ2019214}, such that

% \begin{equation}
% \label{eq:mf}
% X\approx UV,
% \end{equation}
% where $U \in \bm{R^{M\times H}}$,$V \in \bm{R^{H\times N}}$, and normally $H \ll min(M,N)$. 

% The purpose of NMF is to learn the matrices $U$ and $V$ from the matrix $X$. Consequently, the Euclidean distance between $X$ and $UV$ is adopted as loss function to find optimal $U$ and $V$. Therefore the objective is to minimize the loss function expressed as followed. \textcolor{blue}{Also, the Stochastic Gradient Descent(SGD) is adopted to learn the $U$ and $V$, which is described as Table \ref{sgd}.} 

% % \begin{equation}
% % \argmin_{U,V
% % }{\parallel X-UV \parallel}^2
% % \end{equation}
% \textcolor{blue}{
% \begin{equation}
% E(U,V)=\sum{\parallel X_{i,j}-U_{i,:}\cdot V_{:,j}^{T}\parallel}^2
% \end{equation}
% }
% The classical NMF is a reliable method to fill up the sparse matrix due to the hidden vectors make full use of observed data to mining the hidden features. 
% But for tunnel structure located in complicated conditions, only use NMF without considering stress boundary conditions on site is not reasonable. Thus, the distribution characteristics of external load on tunnel structure is employed and integrated into loss function.
This section described the method based on Non-negative Matrix Factorization (NMF) to fill up the constructed sparse matrix. Given the sparse matrix $X \in \bm{R^{M \times N}}$, NMF approximates it with two low-rank matrix $U$ and $V$\cite{TENG2019205,LUDENACHOEZ2019214}, such that

\begin{equation}
\label{eq:mf}
X\approx UV,
\end{equation}
where $U \in \bm{R^{M\times H}}$,$V \in \bm{R^{H\times N}}$, and normally $H \ll min(M,N)$. The resulting of UV is a dense matrix. The similarity between original sparse matrix and resulting dense matrix is limited to those cells that are not empty in the original matrix. And the Stochastic Gradient Descent(SGD) is adopted to learn the $U$ and $V$, which is described as Table \ref{sgd}. 

% The purpose of NMF is to learn the matrices $U$ and $V$ from the matrix $X$. Consequently, the Euclidean distance between $X$ and $UV$ is adopted as loss function to find optimal $U$ and $V$. Therefore the objective is to minimize the loss function expressed as followed. \textcolor{blue}{Also, the Stochastic Gradient Descent(SGD) is adopted to learn the $U$ and $V$, which is described as Table \ref{sgd}.} 

% \begin{equation}
% \argmin_{U,V
% }{\parallel X-UV \parallel}^2
% \end{equation}
NMF has many applications, such as image compression and personalized recommendation. In such application, the image matrix is dense and users’ rating matrix is approximately dense, there is a lot of information that can be used in the $X$, to control overfitting and improve the ability of generalization, L2 regularization is applied in loss function to constrain the matrixs $U$ and $V$. But for tunnel structure,  the matrix $X$ is sparse and the information obtained is less due to the limited number of sensors deployed on a certain section. Furthermore, tunnel structure is located in complicated conditions, the constraints on the matrixs $U$ and $V$ are far different from simple L2 regularization but should correspond to the boundary conditions. Thus, the distribution characteristics of external load on tunnel structure is employed and integrated into loss function as another driving factor.

The Dinghuaimen tunnel is a typical underwater structure, whose overlying layer is mainly soil. The sketch of stress boundary condition is displayed in Figure \ref{fig:soil-water}, in which the water pressure and soil pressure are calculated respectively. Assume that the water level above tunnel section is $h$, and the recommended form of water pressure is expressed as:

\begin{equation}
\label{eq:pw}
P_{w}=\gamma_{w}h
\end{equation}
where $P_{w}$ is water pressure, and $\gamma_{w}$ is the unit weight of water.

% \begin{figure}[H]
%     \centering
%     \includegraphics[width=8cm]{pics/soil-water.png}
%     \caption{Stress boundary condition of underwater tunnel structure.}
%     \label{fig:soil-water}
% \end{figure}

Soil pressure is related to the layers encountered by tunnel section and the locations of load applied. Those loads are named as overburden soil pressure $F_{u}$, lateral soil pressure $F_{s}$, and bottom resistance $F_{b}$, respectively. For each section $i$, the mentioned soil pressure can be calculated as:
\begin{equation}
\label{eq:pw}
F_{u}^{i}=\sum_{j=1}^{n}{h_{j}(\gamma_{j}^{s}-\gamma_{w})},
\end{equation}

\begin{equation}
\label{eq:pw}
F_{b}^{i}=F_{u}^{i}+\frac{G-F}{d},
\end{equation}

\begin{equation}
\label{eq:pw}
F_{s}^{i}=\lambda_{j}{[F_{u}^{i}+(\frac{d}{2}-y)(\gamma_{j}^{s}-\gamma_{w})]},
\end{equation}
where $n$ is the number of stratum encountered by tunnel section. $\lambda_{j} $ is the parameter related to ground resistance. $\gamma_{j}^{s}$ is unite weight of soil.
$d$ is external diameter of tunnel section. $F$ and $G$ are floatage and gravity, respectively. $y$ is the vertical height when regarding the center of tunnel circle as origin of coordinates.

It is obvious that both the geometric construction and the external load applied on tunnel are axisymmetric.Assuming the tunnel structure is an elastic body and the effect of segment joints is ignored, the property of force distribution is also axisymmetric according to the theory of elastic mechanics. Therefore, only right semi-circle of tunnel structure is selected for mechanical analysis. The tunnel lining is divided into infinitesimal units firstly, and the load applied on any unit of upper semi-circle is displayed as Figure \ref{fig:top} shown. 

% \begin{figure}[H]
%     \centering
%     \includegraphics[width=5cm]{pics/top semi-circle.png}
%     \caption{Stress distribution on the top semi-circle of tunnel lining.}
%     \label{fig:top}
% \end{figure}

It can be seen from the schematic diagram that each unit is subjected to four forces in total, namely the lateral soil pressure in horizontal direction, water pressure in radial direction, gravity and overburden soil pressure in vertical direction. If they are decomposed in a same polar coordination, the radial resultant force can be calculated as follows:
\begin{equation}
\label{eq:top sigma}
F_{r}=P_{w}+(G+F_{u})sin\theta+F_{s}cos\theta,
\end{equation}
where $\theta$ is the angel between tangent line and vertical direction.

In addition, the resultant force in tangential direction is expressed as:

\begin{equation}
\label{eq:top tau}
F_{\tau}=F_{s}sin\theta-(G+F_{u})cos\theta,
\end{equation}

% \begin{figure}[H]
%     \centering
%     \includegraphics[width=5cm]{pics/forward semi-circle.png}
%     \caption{Stress distribution on the forward semi-circle of tunnel lining.}
%     \label{fig:forward}
% \end{figure}

Similarly, there are also four force applied on the lower semi-circle shown in Figure \ref{fig:forward}, including the lateral soil pressure in horizontal direction, water pressure in radial direction, gravity and bottom resistance in vertical direction. The resultant forces of radical and tangential direction can be expressed as:

\begin{equation}
\label{eq:forward sigma}
F_{r}=P_{w}+(F_{b}-G)sin\theta+F_{s}cos\theta,
\end{equation}

\begin{equation}
\label{eq:forward tau}
F_{\tau}=F_{s}sin\theta-(G-F_{b})cos\theta,
\end{equation}
where $F_{r}$, $F_{\tau}$ represent the resultant force in radical and tangential direction, respectively.

%Above mechanical analysis of tunnel structure is regarded as constraint conditions to be employed in the loss function of spatial deduction algorithm to guide the factorization process to find $U$ and $V$.% 

The mechanical analysis of tunnel structure captures the distribution characteristics of external load on tunnel structure. These mechanical properties should be employed to guide the factorization process to find $U$ and $V$. Considering that the monitoring section is a continuum body, the force exerted on the tunnel face gradually spreads resulting adjacent area on the tunnel has similar mechanical behaviors, so the information between adjacent points embodied by theoretical formula can be used in the process of factorization. Furthermore, the characteristics of axial symmetry of tunnel could also be employed to guide the process to learn $U$ and $V$. Specifically, two constraints were added to the traditional loss function of matrix factorization. One is to consider the similarity between adjacent points, and the other is to consider the similarity between axisymmetric points. Therefore, the loss function of spatial deduction $E(U,V,\lambda_1,\lambda_2)$ is defined as follows:

\begin{equation}
\begin{split}
\label{eq:mf with domain loss}
E(U,V,\lambda_1,\lambda_2)=\sum_{(i,j)\in\mathbb{A}}{\parallel X_{i,j}-U_{i,:}\cdot V_{:,j}^{T}\parallel}^2 \\ +\lambda_1 \sum_{m=1}^{M}\sum_{n=1}^{N-1}Q_{m,n}{\parallel U_{m,:}\cdot V_{:,n}^{T} - U_{m,:}\cdot V_{:,n+1}^{T}\parallel}^2 \\ +\lambda_2 \sum_{m=1}^{M}\sum_{n=1}^{N}{\parallel U_{m,:}\cdot V_{:,n}^{T} - U_{m,:}\cdot V_{:,N+1-n}^{T}\parallel}^2,
\end{split}
\end{equation}
\begin{equation}
Q_{m,n}=1-\frac{max(\mid F_{\tau_{m,n}} \mid,\mid F_{\tau_{m,n+1}} \mid)-min(\mid F_{\tau_{m,n}} \mid,\mid F_{\tau_{m,n+1}} \mid)}{max(\mid F_{\tau_{m,n}} \mid,\mid F_{\tau_{m,n+1}} \mid)},
\end{equation}
% where $Q_{m,n}\in(0,1)$, $\mathbb{A}$ is the subscript set of non-empty units in the matrix $X$, $\cdot$ denotes the inner product, $U_{i,:}$ is the $i^{th}$ row of matrix $U$, $V_{:,j}^{T}$ is the transpose of  $j^{th}$ coloumn of matrix $V$.
% $M$ is the number of layers split in the tunnel face, $N$ is the number of parts split in each layer. 
% $\lambda_1$ and $\lambda_2$ are the hyper-parameters to  control the extent of the similarity of adjacent areas and symmetrical areas in numerical value respectively. The combination of $\lambda_1$ and $Q_{m,n}$ enables the method to control the extent of the similarity between adjacent areas according to the position in tunnel face guided by theoretical formula. 
where $\sum_{(i,j)\in\mathbb{A}}{\parallel X_{i,j}-U_{i,:}\cdot V_{:,j}^{T}\parallel}^2$ is the loss caused by correctly predicting the monitoring data; $\sum_{m=1}^{M}\sum_{n=1}^{N-1}Q_{m,n}{\parallel U_{m,:}\cdot V_{:,n}^{T} - U_{m,:}\cdot V_{:,n+1}^{T}\parallel}^2$ is the loss to account for the similarity between two adjacent areas; $\sum_{m=1}^{M}\\\sum_{n=1}^{N}{\parallel U_{m,:}\cdot V_{:,n}^{T} - U_{m,:}\cdot V_{:,N+1-n}^{T}\parallel}^2$ is the loss to account for the similarity between two  axisymmetric areas. $\mathbb{A}$ is the subscript set of non-empty units in the matrix $X$, $\cdot$ denotes the inner product, $U_{i,:}$ is the $i^{th}$ row of matrix $U$, $V_{:,j}^{T}$ is the transpose of  $j^{th}$ coloumn of matrix $V$. $\lambda_1$ and $\lambda_2$ are the hyper-parameters to balance the weights between two constraints. $Q_{m,n}$ is a dynamic parameter calculated using the mechanical formula. If the force values of two adjacent areas are more similar in the mechanical formula, the corresponding $Q_{m,n}$ will be larger. On the contrary, if the force values of the two adjacent areas differ in the mechanical formula, the corresponding $Q_{m,n}$ will be smaller. Combining $\lambda_1$ and $Q_{m,n}$ enables this constraint to more accurately adjust the similarity between two adjacent areas according to the mechanical formula which makes the result more reasonable.
With two constraints, the proposed method fills up the empty units with the structural characteristics information of the tunnel.

\subsection{Evaluation metrics for deduction results}

Three evaluation metrics are adopted to measure the effectiveness of spatial deduction model: Root Mean Square Error (RMSE), Mean Absolute Error (MAE) and Pearson Correlation Coefficient (PCC). 
% A lower value is better for RMSE and MAE, a high value is better for PCC.

RMSE is used to measure the deviation between the prediction value and the true value. 
% The magnitude of RMSE varies from 0 to $+\infty$. 
It is calculated as:

\begin{equation}
    RMSE=\sqrt{\frac{1}{n}\sum_{i=1}^{n}{(y_i-\hat{y_i
    })}^2}
\end{equation}

MAE is the mean of absolute error between the prediction value and the true value which can better reflect the actual situation of the prediction value error.
% The magnitude of MAE varies from 0 to $+\infty$. 
It is calculated as:

\begin{equation}
    MAE=\frac{1}{n}\sum_{i=1}^{n}{\mid y_i-\hat{y_i
    } \mid }
\end{equation}

PCC describes the linear correlation between two variables. The magnitude of PCC varies from -1 to 1, where 1 means total positive linear correlation, 0 means no linear correlation, and -1 means total negative linear correlation.It is calculated as: %When two variables are covariant, PCC is greater than 0, when contravariant, PCC is less than 0, when linearly independent, PCC is equal to 0.% 

\begin{equation}
    PCC=\frac{\sum\nolimits_{i=1}^{n}(y_i-\bar{y_i})(\hat{y_i}-\bar{\hat{y_i}})}{\sqrt{\sum\nolimits_{i=1}^{n}{(y_i-\bar{y_i})}^2 \sum\nolimits_{i=1}^{n}{(\hat{y_i}-\bar{\hat{y_i}})}^2 }}
\end{equation}
%where $y_i, \hat{y_i}$ and $n$ are ground truth, the prediction and the size of the dataset respectively. 
where $y_i$ is ground truth, $\hat{y_i}$ is prediction value, $n$ is the size of the dataset.

%% file: 4-Expriment.tex
\section{Data experiment and discussion of deduction results}
\label{S:4}

This section aims to demonstrate the superiority of the  proposed spatial deduction algorithm on the monitoring dataset obtained from SHMS of Dinghuaimen tunnel. To better verify the presented model, the cross-test is adopt by changing the testing set which consists of the monitoring data collected from different positions. Also, the differences between deduction results and actual conditions are elaborately discussed. The source code is available\footnote[1]{https://github.com/wyhhuster/Spatial-deduction}.

\subsection{Dataset information obtained from SHM}
Considering the total number of monitoring sections and sensors are large, three typical monitoring sections located at different geological conditions are selected for testing experiment. The number of these sections are S2, S4, S9. Specifically, S2 is under land connection portion, S4 is underwater and cross multiple layers, and S9 is also underwater but only cross one layer. The number of stress sensors in working order is 9, 15, and 6, respectively. These sensors record mechanical behaviors once a day, totally 436 days. In order to avoid the effect of initial value, the data are relative variation of initial value. The geological conditions of these three sections and the property parameters are displayed in Figure \ref{fig:S4} and Table \ref{tab:s4}.

% \begin{figure}[H]
%     \centering
%     \includegraphics[height=6.3cm]{pics/three.png}
%     \caption{Geological conditions of some typical monitoring sections: (a) S2, (b) S4, and (c) S9.}
%     \label{fig:S4}
% \end{figure}

% \begin{table}[H]
%     \centering
%     \begin{tabular}{l c c}
%     \hline
%     \textbf{Ground types} & \textbf{Lateral pressure coefficient} & \textbf{Unite weight}
%     \\
%     \hline
%     Silt & 0.43 & 19.4\\
%     Fine sand & 0.40 & 19.3\\
%     Silt clay & 0.65 & 18.6\\
%     Gravel & 0.25 & 20.6\\
%     Gravel sand & 0.30 & 20.3\\
%     Weather siltstone & 0.14 & 19.2\\
%     \hline
%     \end{tabular}
%     \caption{Mechanical parameters of surrounding rock}
%     \label{tab:s4}
% \end{table}

 %\textbf{Ground types} & \textbf{$\lambda_{j}$} &  \textbf{$\gamma_{j}^{s}$}

\subsection{Formalization on real-time monitoring data of tunnel structure}
The monitoring scheme has demonstrated that each ring of tunnel lining consists of 10 segments and stress sensors are installed in inner and outer of tunnel lining. To maximize the use of observed monitoring data, the median value is represented by the average of the monitoring data from the internal and external sensors. The corresponding reconstruction of tunnel model is shown as Figure \ref{fig:partition}. The model of monitoring section is divided into $M=3$ layers, inner layer corresponds to the inner tunnel lining, middle layer corresponds to the middle tunnel lining and outer layer corresponds to the outer tunnel lining. Furthermore, each layer is divided into $N=50$ parts with each segment has total 150 parts. The naming and split for each segment is marked in Figure \ref{fig:partition} and the parts colored in yellow are where the sensors have been deployed in segments. The yellow parts is named with the segment name where it is located as the prefix, and the layer where it is located as the suffix, such as B1-inner means this yellow part is located in B1 segment and located in inner layer.
%
%The yellow parts that are in the same position of their layers are named with the same prefix and differ from each with the different suffix: inner, middle, outer, such as B1-inner, L2-outer.
In S2, the observed data is obtained directly from 13 parts. Flattening out the split $S2$ section model, a matrix with 3 rows and 50 columns will be got where there are 13 units are non-empty. $S4$ and $S9$ have the same number of rows and columns, but 22 units and 8 units are non-empty respectively. 
Based on the field investigation results shown in Figure \ref{fig:S4} and Table \ref{tab:s4}, the mechanical analysis of these sections are calculated according to equations  \ref{eq:pw}-\ref{eq:forward tau}, and then employed in equation 12. Given $\mathbb{A}$ represents the subscript set of non-empty element in matrix, the non-empty elements of different sections are expressed as:

$\mathbb{A}$ for $S2$ is $\{(1,2),(1,8),(1,14),(1,42),(1,48),(2,2),(2,8),(2,42),(2,48),\\(3,2),(3,8),(3,42),(3,48) \}$,

$\mathbb{A}$ for $S4$ is $\{(1,2),(1,8),(1,23),(1,28),(1,37),(1,33),(1,42),(1,46),(2,2),\\(2,8),(2,23),(2,37),(2,33),(2,42),(2,46),(3,2),(3,8),(3,23),(3,37),(3,33),\\(3,42),(3,46)\}$,  

$\mathbb{A}$ for $S9$ is $\{(1,4),(1,13),(2,4),(2,13),(3,4),(3,13),(3,19),(3,33)\}$
%Thus, the loss function used for the reconstruction of $S4$ monitoring section is expressed as follows.

% \begin{equation}
% \begin{split}
% \label{eq:mf with domain loss}
% E(U,V,\lambda_1,\lambda_2)=\sum_{(i,j)\in\mathbb{A}}{\parallel X_{i,j}-U_{i,:}\cdot V_{:,j}^{T}\parallel}^2 \\ +\lambda_1 \sum_{m=1}^{3}\sum_{n=1}^{49}Q_{m,n}{\parallel U_{m,:}\cdot V_{:,n}^{T} - U_{m,:}\cdot V_{:,n+1}^{T}\parallel}^2 \\ +\lambda_2 \sum_{m=1}^{3}\sum_{n=1}^{50}{\parallel U_{m,:}\cdot V_{:,n}^{T} - U_{m,:}\cdot V_{:,51-n}^{T}\parallel}^2 
% \end{split}
% \end{equation}

% \begin{figure}[H]
% \centering\includegraphics[width=14cm]{pics/S249.png}
% \caption{Partition form of monitoring sections.}
% \label{fig:partition}
% \end{figure}

\subsection{Crossing experiment with various testing points}

To verify the reliability of presented spatial deduction model, three independent experiments are conducted by changing testing datasets in every sections. So there are total 9 experiments. Process for each experiment is shown in Figure \ref{fig:evaluation}. In every experiment, one point is randomly selected from the non-empty points as testing point, which means the stress value of the selected point will not be used for model training, and the stress value of the selected point is testing dataset. The resting points are training points and their stress values are training dataset. Then grid search and K-fold cross validation are employed on training dataset to find the best hyper-parameters $\lambda_1$ and $\lambda_2$. Grid search and K-fold cross-validation are usually used together to find the optimal hyper-parameters. Grid search enumerates the possible values of each hyper-parameter, different values of different hyper-parameters are combined with each other 
which forms the possible value pairs for all hyper-parameters. For example if the possible values for $\lambda_1$ are 0.1 and 0.2, the possible values for $\lambda_2$ are 0.8 and 0.9, grid search is to enumerate (0.1, 0.8), (0.1, 0.9), (0.2, 0.8), (0.2, 0.9) to find the best combination. Each combination value of $\lambda_1$ and $\lambda_2$ will be evaluated by K-fold cross-validation. After removing the test data from the original data, K-fold cross-validation divides the new data into K parts again, each time using different K-1 parts of data for training, the remaining part of data for validation, the average of K times validation results is the final results. The value pair which has the best validation result is the optimal values for hyper-parameters. Finally the deduction model will use the optimal hyper-parameters to deduct and test on testing dataset. After deduction, three evaluation metrics were used to compare the deduction values at testing points with true values. Totally 9 monitoring points were selected as testing points to conduct experiment respectively, S2-B2-inner, S2-F-middle, S2-L1-outer, S4-B1-inner, S4-F-middle, S4-B3-outer, S9-F-inner, S9-B7-middle, S9-L1-outer. In the meantime, the subscripts corresponding with test points should be removed from the set of subscript $\mathbb{A}$ which are (1, 8), (2, 42), (3, 48) for $S2$ section ,(1, 47), (2, 37), (3, 8) for $S4$ section, (1, 13), (2, 4), (3, 19) for $S9$ section respectively. Therefore, there are 12, 21, 7 monitoring points can be used to train the process of spatial deduction in each section. The values of two hyper-parameters are obtained  from a 6-fold cross-validation process with $\lambda_1$ ranges in $\{0.1,0.2,...,1\}$ and $\lambda_2$ ranges in $\{0.1,0.2,...,1\}$ for $S2$, 7-fold cross-validation process with $\lambda_1$ ranges in $\{0.1,0.2,...,1\}$ and $\lambda_2$ ranges in $\{0.1,0.2,...,1\}$ for $S4$, 7-fold cross-validation process with $\lambda_1$ ranges in $\{0.1,0.2,...,1\}$ and $\lambda_2$ ranges in $\{0.1,0.2,...,1\}$ for $S9$. The values of $\lambda_1$ are all 0.1 and the values of $\lambda_2$ are all 0.1 
in nine experiments. Early stopping strategy is applied in the train process to avoid overfitting. In addition, to benchmark the performance of the presented method, the same experiments are conducted on baseline method using original NMF without boundary condition.

\subsection{Discussion for spatial deduction result}

% In this section, we compare the deduction value and actual monitoring value on test points of nine experiments. Figure \ref{fig:compare} shows the 436 days of actual monitoring value and the deduction value on nine test points (each section has three points) respectively. It is obvious that the deduction values agree well with actual monitoring results, which indicates the proposed method can capture the structure mechanical information to deduct the mechanical behaviors where no sensors have been deployed. Table 2 shows the evaluation results of RMSE, MAE and PCC, which denotes that the proposed method achieves a good performance.
In this section, the experiment results conducted by our presented method are compared with other baseline. Figure \ref{fig:compare} shows the 436 days of actual monitoring value and the deduction values on 9 test points (each section has three points) respectively. It is obvious that the deduction values using our presented method agree well with actual monitoring results, which indicates the proposed method can capture the structure mechanical information to deduct the mechanical behaviors where no sensors have been deployed. Table \ref{tab:res} shows the evaluation results of RMSE, MAE and PCC, which denotes that the proposed method achieves a good performance. It can also found that the baseline performs worse especially in the experiments on S9 section. This is because the useful information in S9 section is very limited and gradient information is not transmitted through  units. For example, when selecting the S9-L1-outer as test point, there is no gradient information in neither L1-middle nor L1-inner. The gradient information from F-outer and B3 outer can not be passed to L1-outer because there is no guidance form  boundary condition, so there is no reasonable results. The proposed method takes into account the boundary conditions. The two constraints in equations \ref{eq:mf with domain loss} enables the gradient information pass in adjacent areas making all untis have gradient information.

%% file: 5-Application.tex
\section{Application:tunnel full-faced behaviors at different time scales}
\label{S:5}
As an important application, the proposed spatial deduction algorithm is employed to conduct the full face mechanical behaviors of Dinghuaimen tunnel, including the characteristics of current stress distribution and historical performance. 

\subsection{Deduction for current stress distribution of tunnel structure}
It has been proved that the presented spatial deduction model is reliable to deduce the unknown state with a few observed  monitoring data. In this section, this model is further verified to deduce the current mechanical behaviors of tunnel full face. Different to the conducted experiments, all of the monitoring points are used for model training, while the other positions with no sensors monitoring are to deduct. The current monitoring data of inner and outer sensors, and the calculated results of middle layers are displayed in Figure \ref{fig:appli}. Based on these observed data information, the characteristics of current stress distribution are deduced and expressed through cloud picture, as shown in Figure \ref{fig:dec}. It can be found the stress on inner lining is larger than that on outside for these three sections. The maximum value of relative stress variation is located 
between the arch bottom and hance, arch bottom and arch crown for $S2$, $S4$, $S9$ respectively. And the values are 18.225KN, 18.380KN, 12.940KN correspondingly.

% \begin{figure}[tbh!]
%     \centering
%     \includegraphics[width=14cm]{pics/applicationthree.jpg}
%     \caption{Current monitoring data of three sections.}
%     \label{fig:appli}
% \end{figure}

% \begin{figure}[tbh!]
%     \centering
%     \includegraphics[width=14cm]{pics/threededu.jpg}
%     \caption{Deduction results on monitoring section.}
%     \label{fig:dec}
% \end{figure}

% \begin{figure}[tbh!]
% \centering\includegraphics[width=14cm]{pics/S2_danger.jpg}
% \caption{The historical performance of six dangerous positions in S2: (a)arch bottom, (b) hance in left-up semicircle, (c) hance in left-low semicircle, (d) hance in right-low semicircle, (e) arch crown,  and (f) hance in right-up semicircle.}
% \label{fig:S2_appli}
% \end{figure}
%\begin{figure}[H]
%\centering
%\subfigure[]{
%\includegraphics[width=6cm]{pics/pic_0.png}
%\caption{fig1}
%}
%\quad
%\subfigure[]{
%\includegraphics[width=6cm]{pics/pic.png}
%}
%\caption{Mechanical behaviors of monitoring deduction: (a) areas of known mechanical behaviors and corresponding values and (b) deduction results on monitoring section}
%\end{figure}

\subsection{Deduction for historical behaviors of tunnel full face}

Not only current stress distribution, the historical performance of unknown points is also deduced via the presented model. 
%The position with large force is dangerous, which can be found through processing spatial deduction results.
In order to evaluate the stability of structure, much more attention should be paid on the positions with large magnitude of stress variation. If these magnitudes do not reach or exceed the warning values, the structure is regarded as stable. Nevertheless, the supporting measures should be implemented as soon as possible. Based on the observed monitoring data in past 436 days, the variation of historical mechanics in tunnel full face is calculated via spatial deduction model. Specifically, six points with large stress value are selected on each section, as shown in Figure \ref{fig:S2_appli},\ref{fig:S4_appli},\ref{fig:S9_appli}. These figures indicated that the variation trend of lining stress is similar among all positions, which varied with seasons obviously. But the amplitude is different, where the large variation amplitude focused on arch crown, hance, and arch bottom. The maximum value of every section is 40.828KN, 43.529KN, 32.362KN respectively. But the corresponding minimum value is 38.727KN, 23.296KN, 13.574KN respectively. Even so, all of the magnitudes do not exceed the warning value of tunnel lining. The tunnel structure is stable. 

%\begin{figure}[H]
%\centering\includegraphics[width=1\linewidth]{pics/danger.png}
%\caption{Locations of top6 dangerous positions}
%\end{figure}

% \begin{figure}[tbh!]
% \centering\includegraphics[width=14cm]{pics/apps_all.png}
% \caption{The historical performance of six dangerous positions in S4: (a)hance, (b) hance in right-up semicircle, (c) arch bottom, (d) arch bottom, (e)hance in left-low semicircle,  and (f) hance in left-up semicircle.}
% \label{fig:S4_appli}
% \end{figure}

% \begin{figure}[tbh!]
% \centering\includegraphics[width=14cm]{pics/S9_danger.jpg}
% \caption{The historical performance of six dangerous positions in S9: (a)hance in right-up semicircle, (b) hance, (c)arch bottom, (d) hance in left-low semicircle, (e) arch bottom,  and (f) arch crown.}
% \label{fig:S9_appli}
% \end{figure}

% \begin{figure}[H]
% \centering
% \subfigure[]{
% \includegraphics[width=6cm]{pics/app1.png}
% %\caption{fig1}
% }
% \quad
% \subfigure[]{
% \includegraphics[width=6cm]{pics/app2.png}
% }
% \quad
% \subfigure[]{
% \includegraphics[width=6cm]{pics/app3.png}
% }
% \quad
% \subfigure[]{
% \includegraphics[width=6cm]{pics/app4.png}
% }
% \quad
% \subfigure[]{
% \includegraphics[width=6cm]{pics/app5.png}
% }
% \quad
% \subfigure[]{
% \includegraphics[width=6cm]{pics/app6.png}
% }
% \caption{The historical performance of six dangerous positions for 436 days: (a)arch bottom, (b) hance in left-up semicircle, (c) hance in left-low semicircle, (d) hance in right-low semicircle, (e) arch crown,  and (f) hance in right-up semicircle.}
% \end{figure}

%% file: 6-Conclusion.tex
\section{Conclusion}
\label{S:6}
In this study, a spatial deduction algorithm was performed to analyze the full face mechanical behaviors of tunnel structure, which integrates actual mechanical properties, real-time monitoring in field into a data-driven model. The main conclusions are summarized as follows.

(1) Different with the previous pure data-driven model, the actual stress boundary conditions are considered and employed in spatial deduction algorithm to guide model training and optimization. Tunnel mechanical model is regarded as axisymmetric, and the characteristics of stress distribution in tunnel lining is discussed in detail, which occupy the key component of loss function to develop novel data-driven model.

(2) In order to verify the reliability of presented model, the cross-test is conducted on the monitoring data collected from Dinghuaimen tunnel. The experiment results denoted that the deduction values agree well with actual conditions, thus the presented model is reasonable.

(3) As a crucial application, the characteristics of current stress distribution and historical performance in tunnel full face are obtained from the deduction results. It is obvious that the stresses located in arch crown, hance, and arch bottom are larger than other positions. The stress on inner lining is larger than that on outside. The maximum value of variation amplitude in 436 days is about 43.431KN and far less than the ultimate strength. 

%% file: 7-Acknowledgement.tex
\section{Acknowledgement}
We thank the reviewers for their constructive comments on this research work. This work is supported by the National Key R\&D Program of China No. 2018YFB2101003, the National Natural Science Foundation of China under Grant No. 51991395, U1806226, 51778033, 51822802, 71901011, U1811463, the Science and Technology Major Project of Beijing under Grant No.\\ Z191100002519012.

%% file: Pics.tex
\section *{Appendix: Figures}
\begin{figure}[H]
    \centering
    \includegraphics[width=12cm]{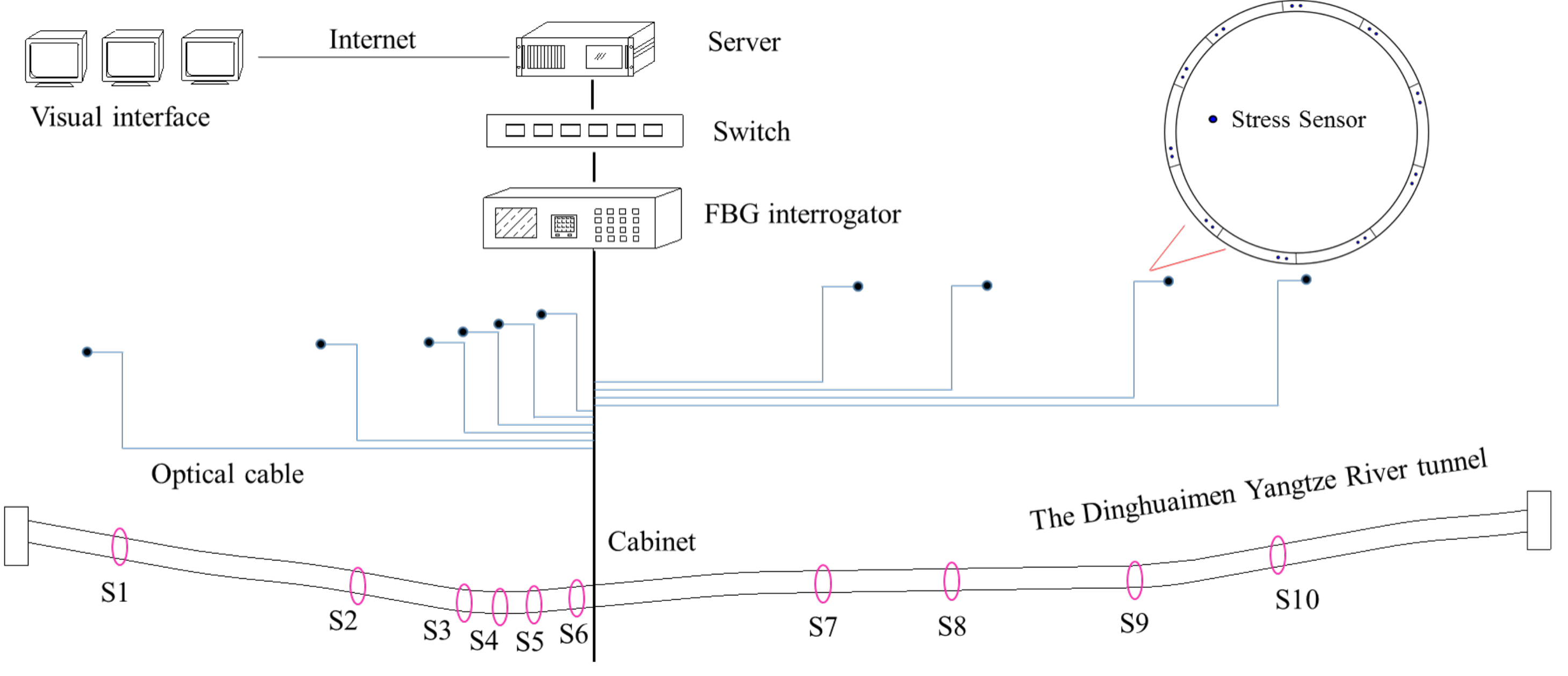}
    \caption{Monitoring scheme and system networking in the study site.}
    \label{fig:layout}
\end{figure}

\clearpage

\begin{figure}[h]
\centering\includegraphics[width=0.7\linewidth]{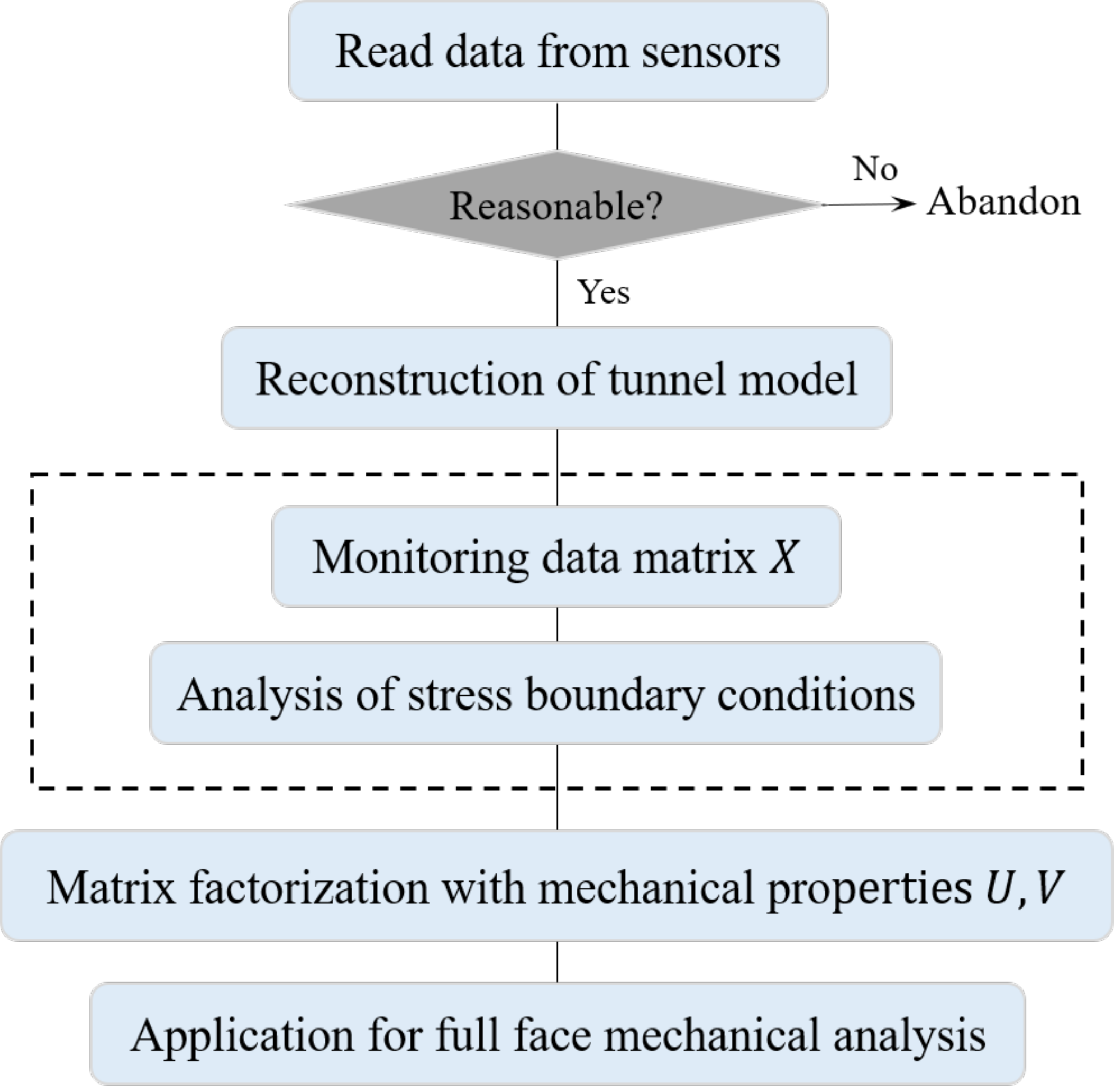}
\caption{Flowchart of the proposed spatial deduction model.}
\label{fig:flowchart}
\end{figure}

\clearpage

\begin{figure}[H] 
    \begin{minipage}[t]{0.5\linewidth}
    \centering 
    \includegraphics[height=4.0cm,width=4.0cm]{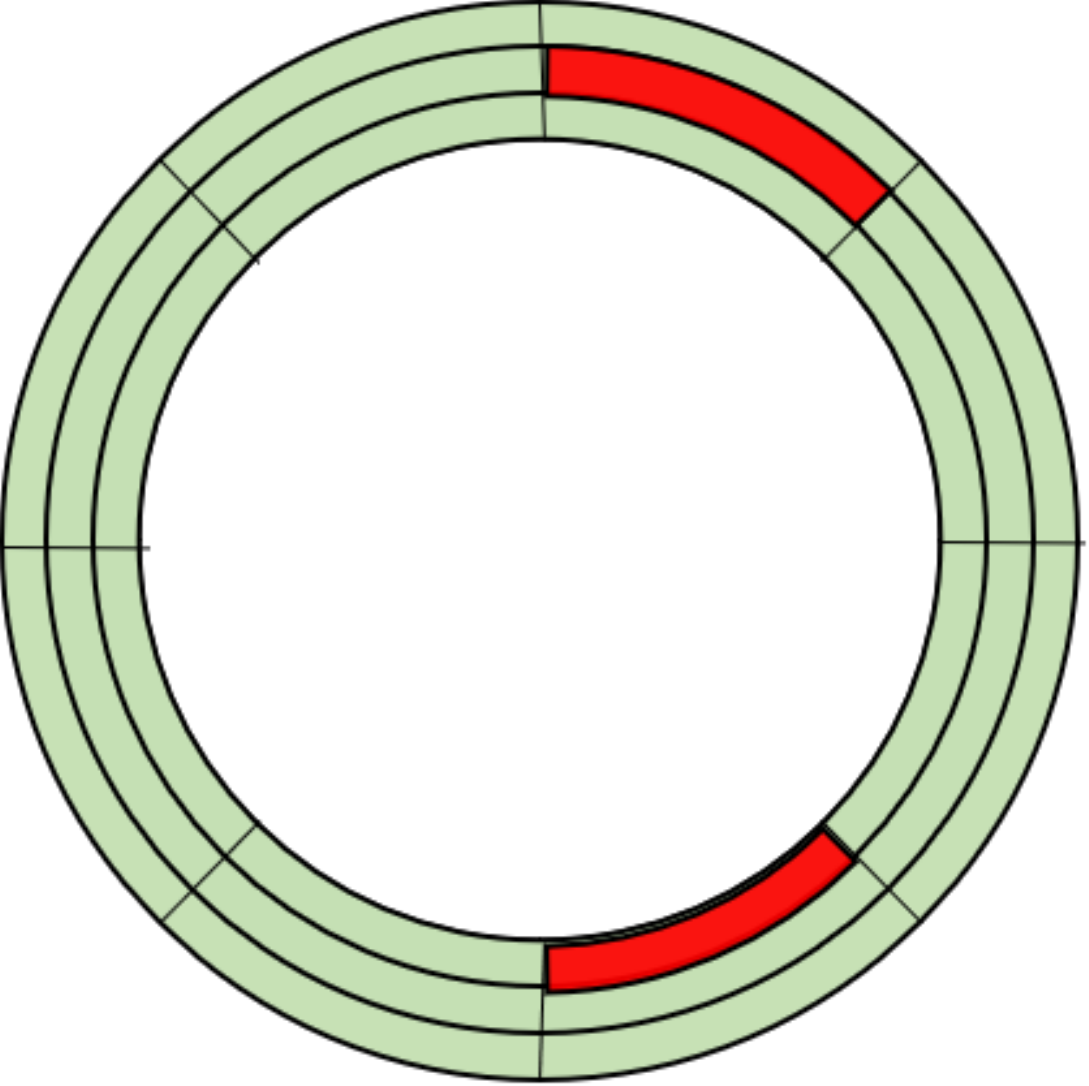} {\label{fig:side:a} }
    \caption*{(a)} 
    \end{minipage} 
    \begin{minipage}[t]{0.5\linewidth}
    \centering 
    \includegraphics[height=2.8cm,width=3.8cm]{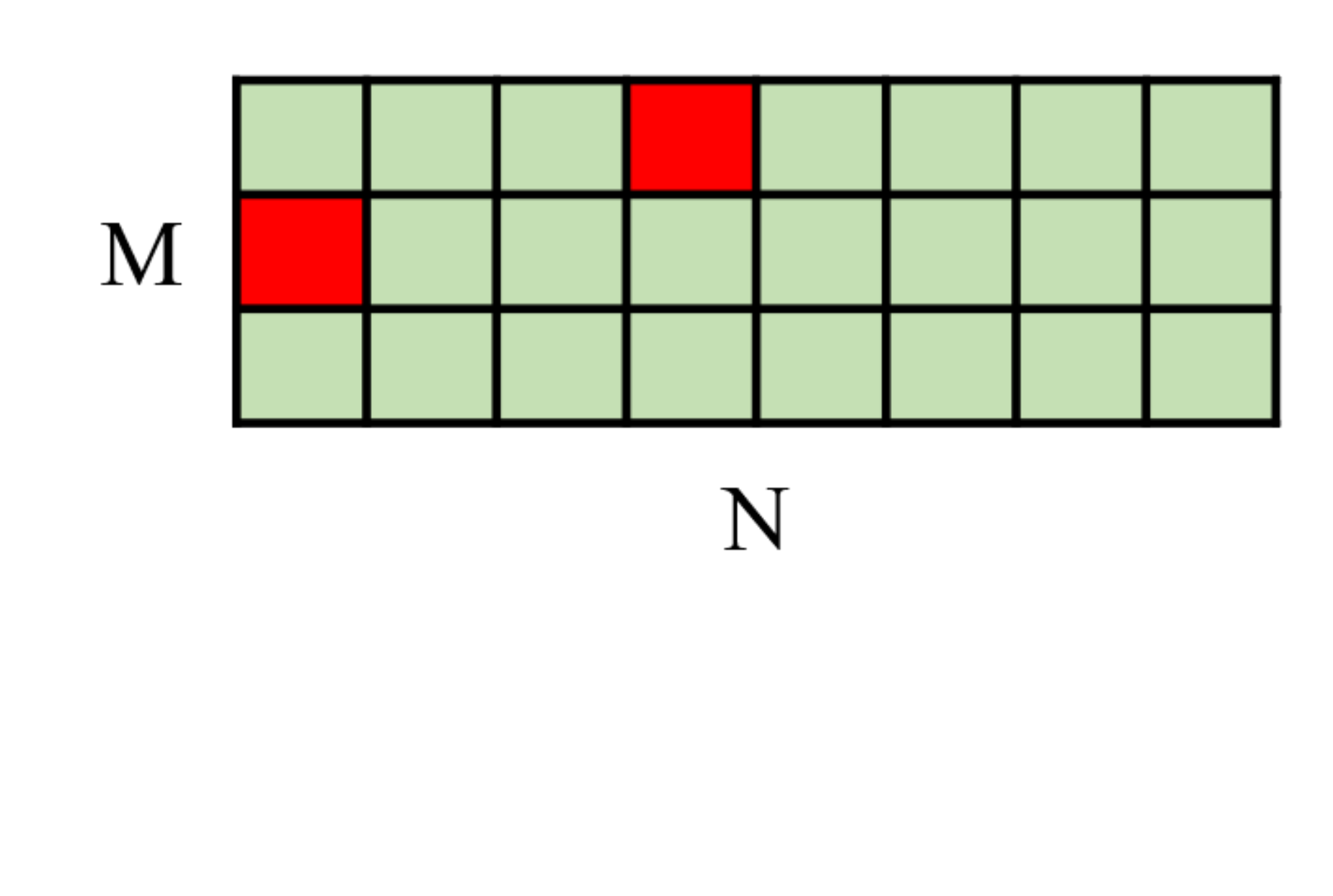}
    \caption*{(b)} 
    \label{fig:side:b} 
    \end{minipage} 
    \caption{The process of reconstructing the tunnel model: (a) division model of formal tunnel, and (b) transform the circle model into a form of matrix.}
\label{fig:reconstruct}
\end{figure} 

\clearpage

\begin{figure}[H]
    \centering
    \includegraphics[width=8cm]{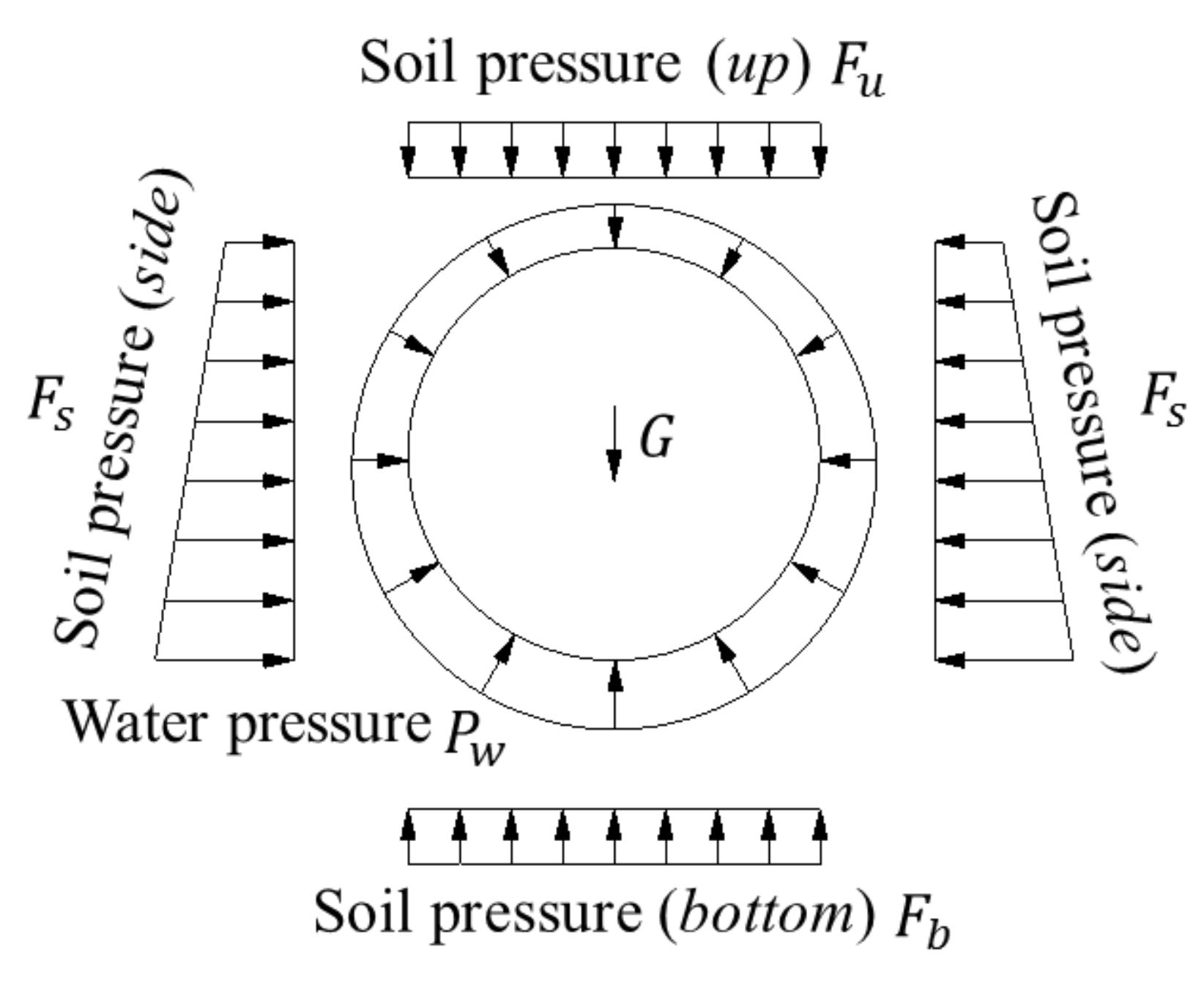}
    \caption{Stress boundary condition of underwater tunnel structure.}
    \label{fig:soil-water}
\end{figure}

\clearpage

\begin{figure}[H]
    \centering
    \includegraphics[width=5cm]{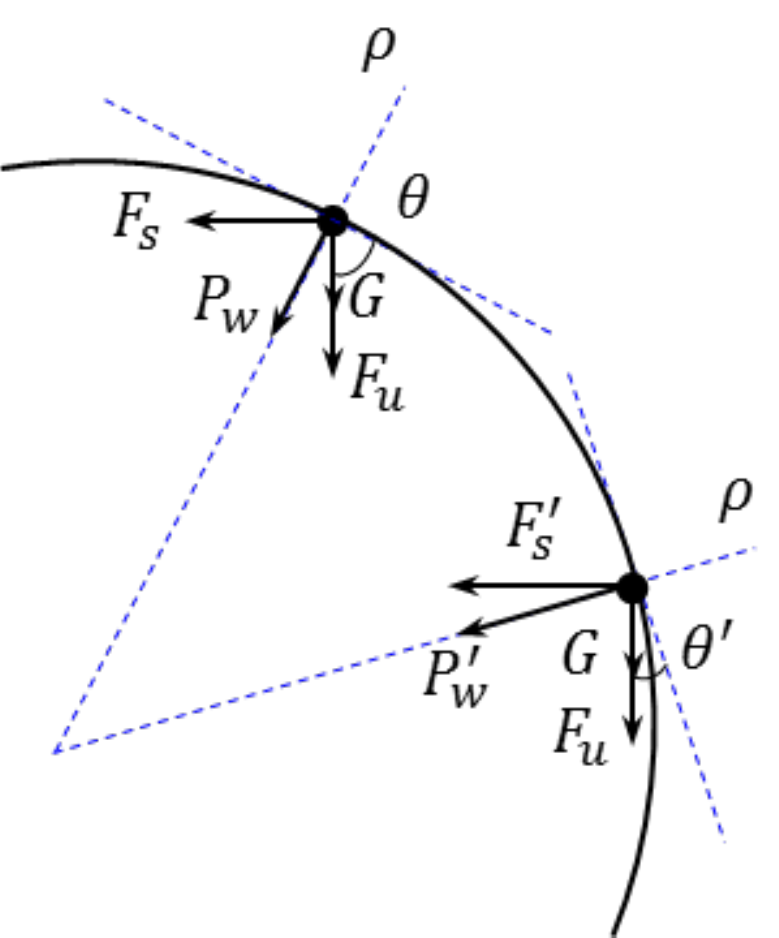}
    \caption{Stress distribution on the top semi-circle of tunnel lining.}
    \label{fig:top}
\end{figure}

\clearpage

\begin{figure}[H]
    \centering
    \includegraphics[width=5cm]{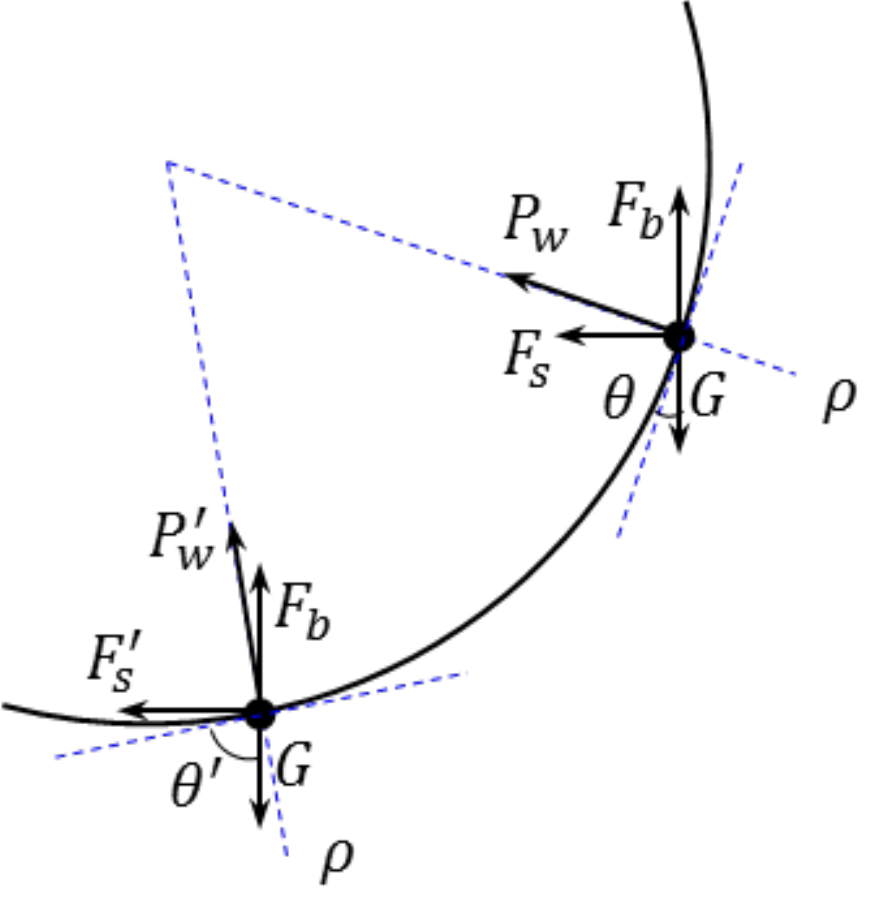}
    \caption{Stress distribution on the forward semi-circle of tunnel lining.}
    \label{fig:forward}
\end{figure}

\clearpage

\begin{figure}[H]
    \centering
    \includegraphics[height=6.3cm]{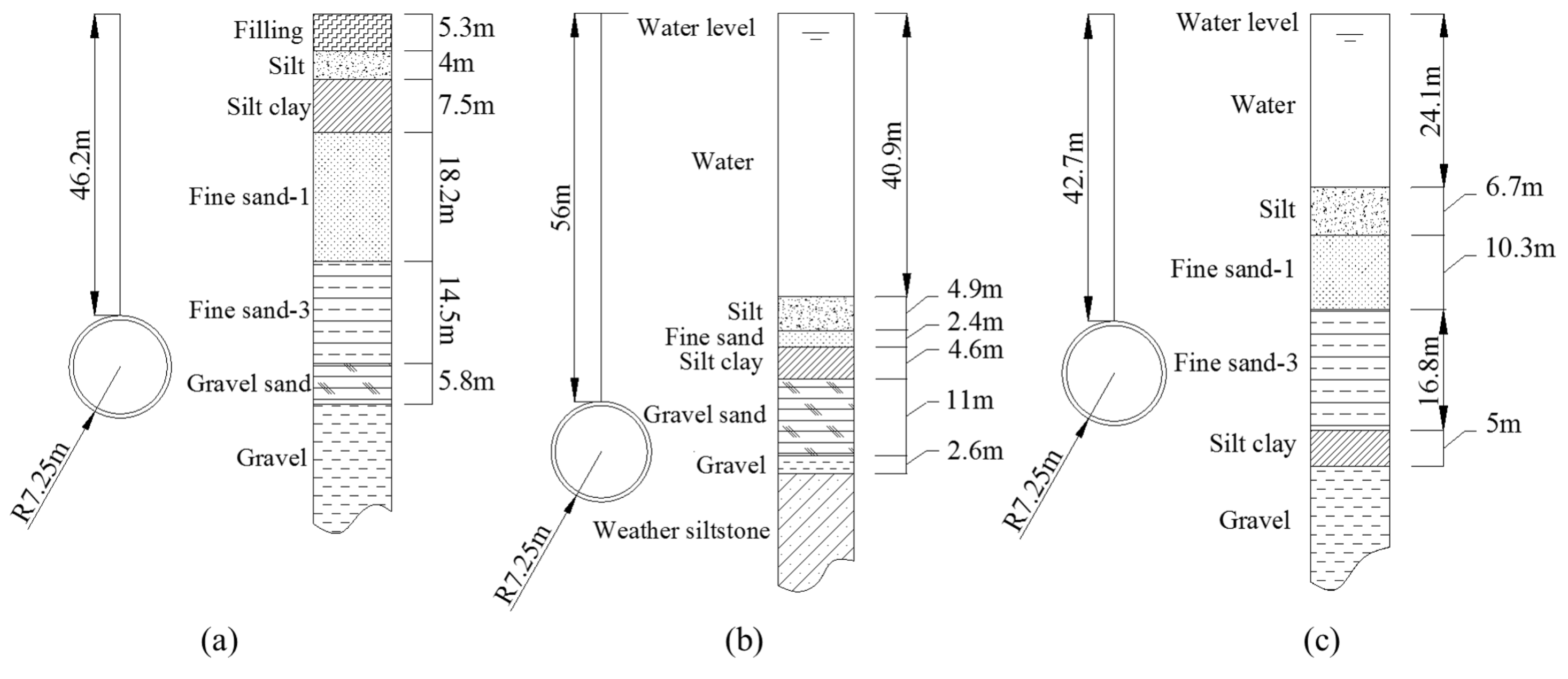}
    \caption{Geological conditions of some typical monitoring sections: (a) S2, (b) S4, and (c) S9.}
    \label{fig:S4}
\end{figure}

\clearpage

\begin{figure}[H]
\centering\includegraphics[width=14cm]{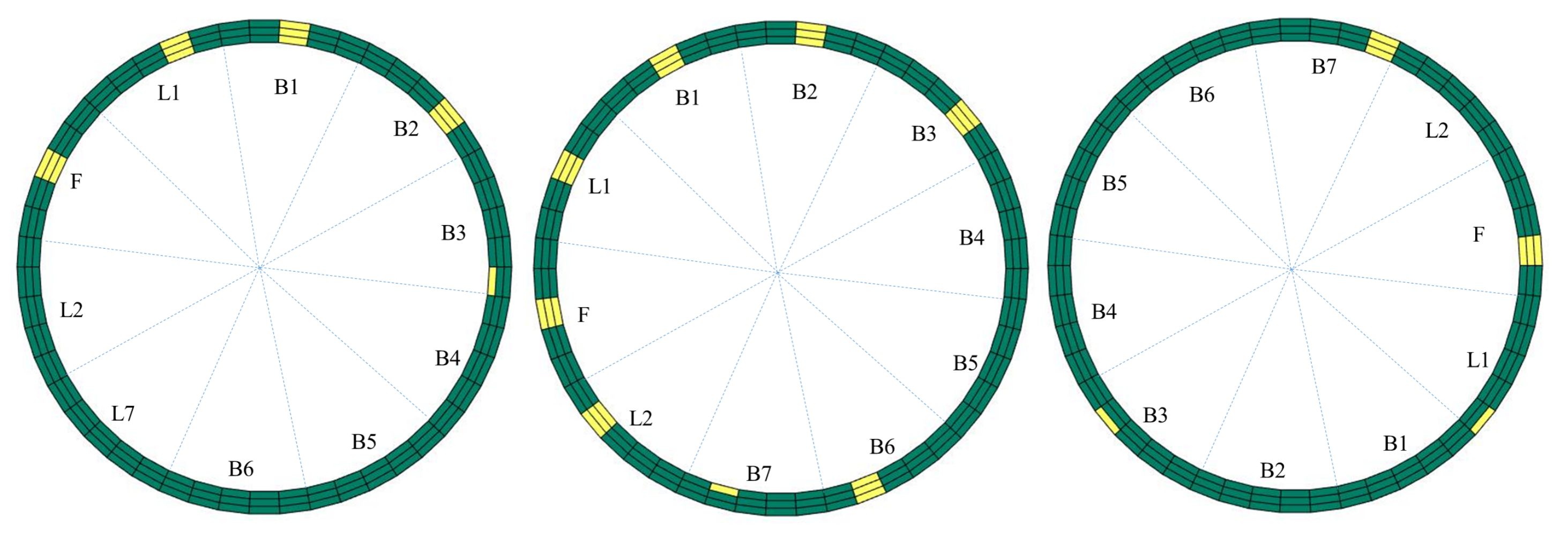}
\caption{Partition form of monitoring sections.}
\label{fig:partition}
\end{figure}

\clearpage

\begin{figure}[H]
    \centering
    \includegraphics[width=8cm]{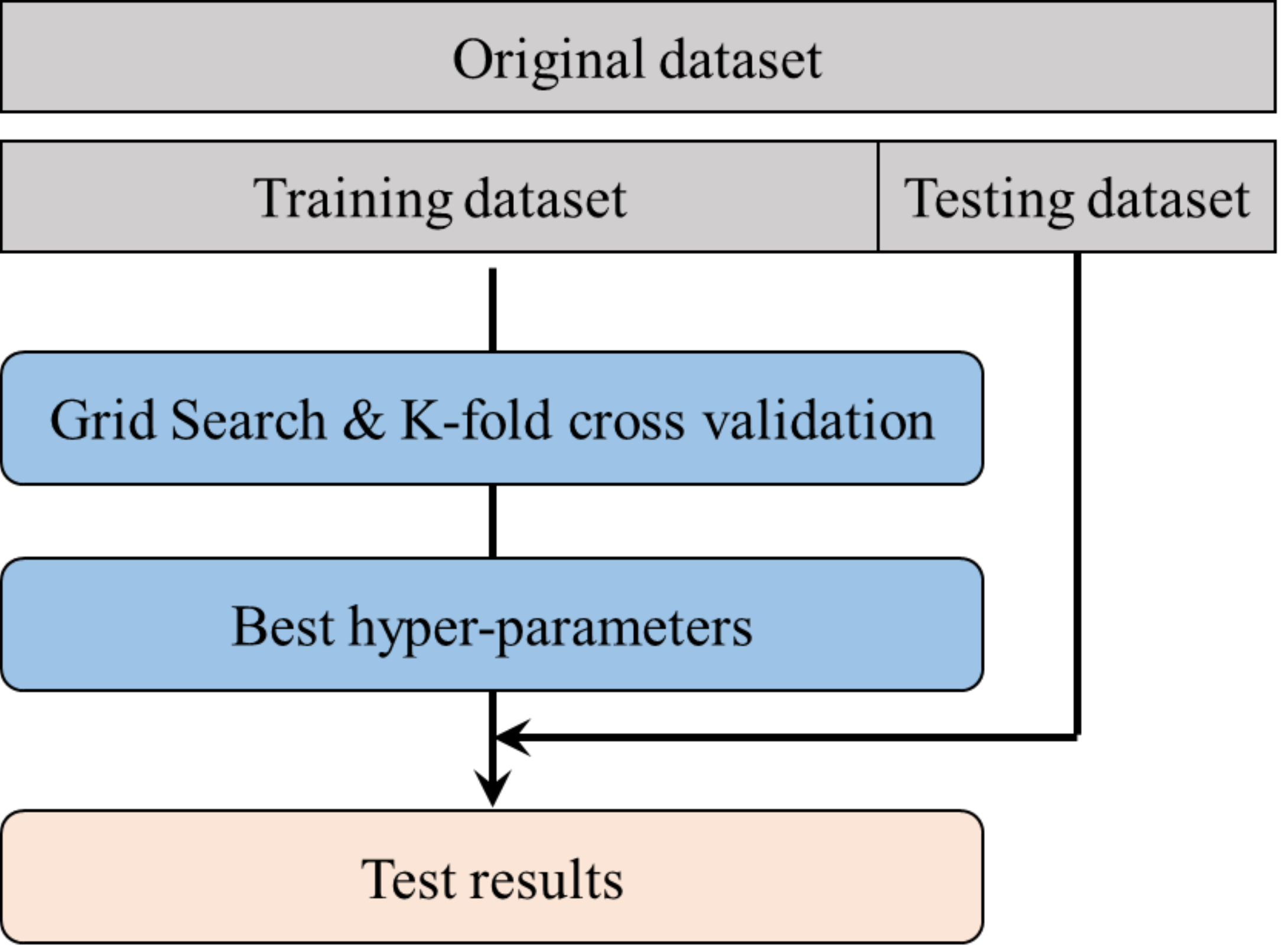}
    \caption{Process of experiment.}
    \label{fig:evaluation}
\end{figure}

\clearpage

\begin{figure}[H]
\centering\includegraphics[width=14cm]{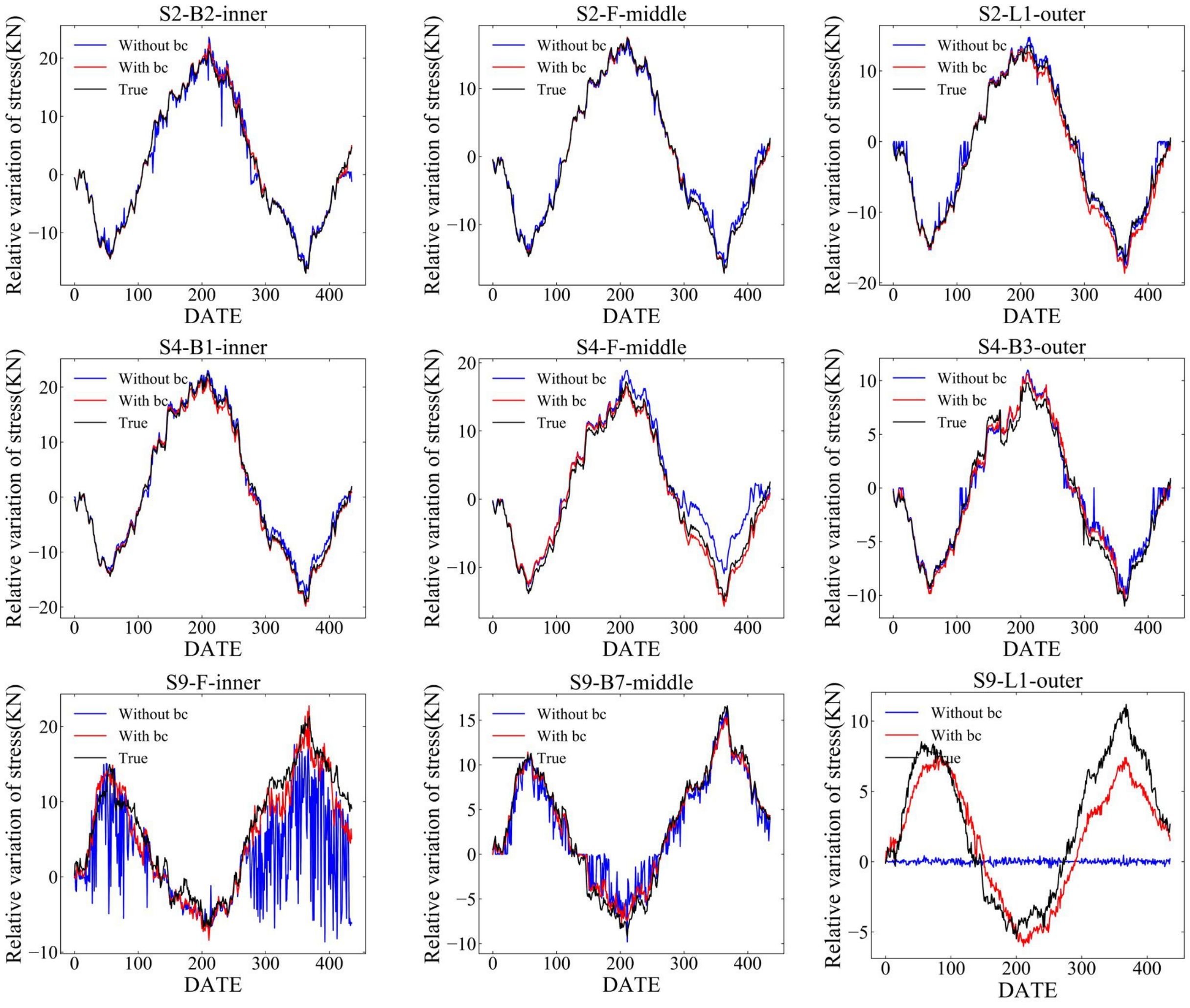}
\caption{Comparison of monitoring and deduction values on test points for 436 days.}
\label{fig:compare}
\end{figure}

\clearpage

\begin{figure}[H]
    \centering
    \includegraphics[width=14cm]{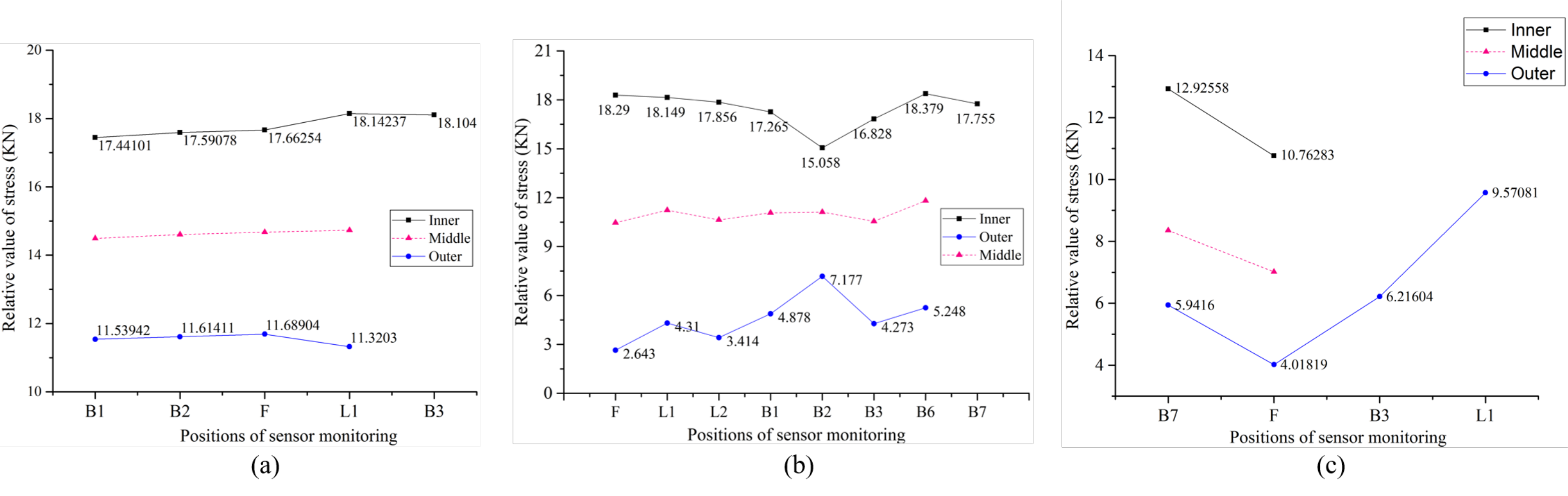}
    \caption{Current monitoring data of three sections.}
    \label{fig:appli}
\end{figure}

\clearpage

\begin{figure}[H]
    \centering
    \includegraphics[width=14cm]{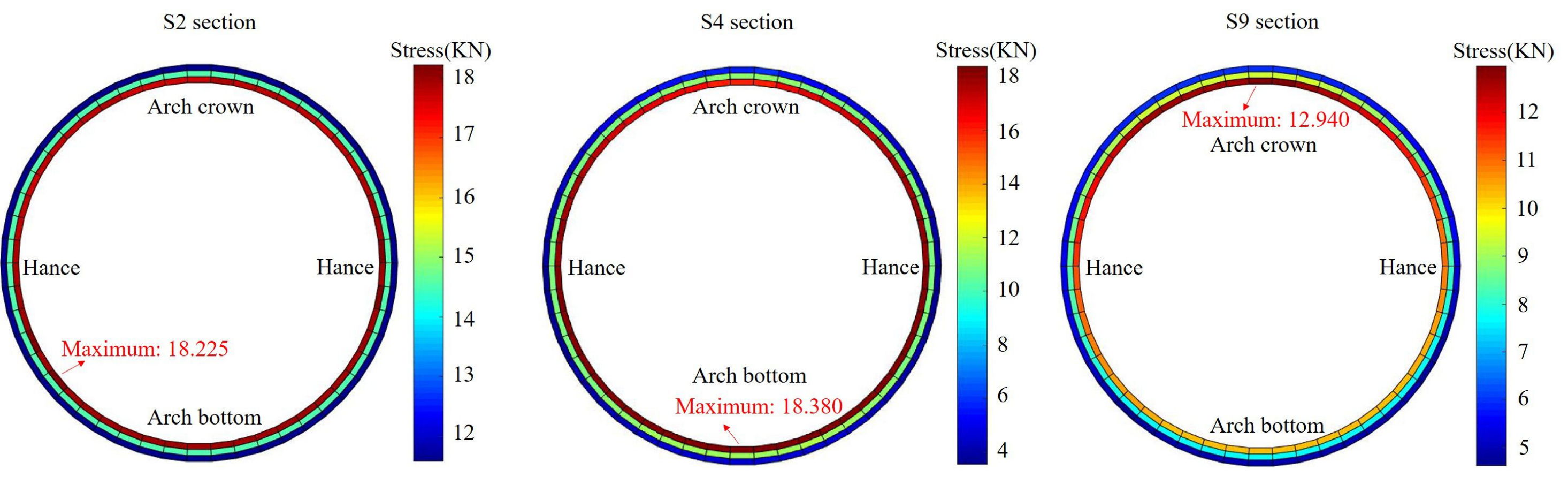}
    \caption{Deduction results on monitoring section.}
    \label{fig:dec}
\end{figure}

\clearpage

\begin{figure}[H]
\centering\includegraphics[width=14cm]{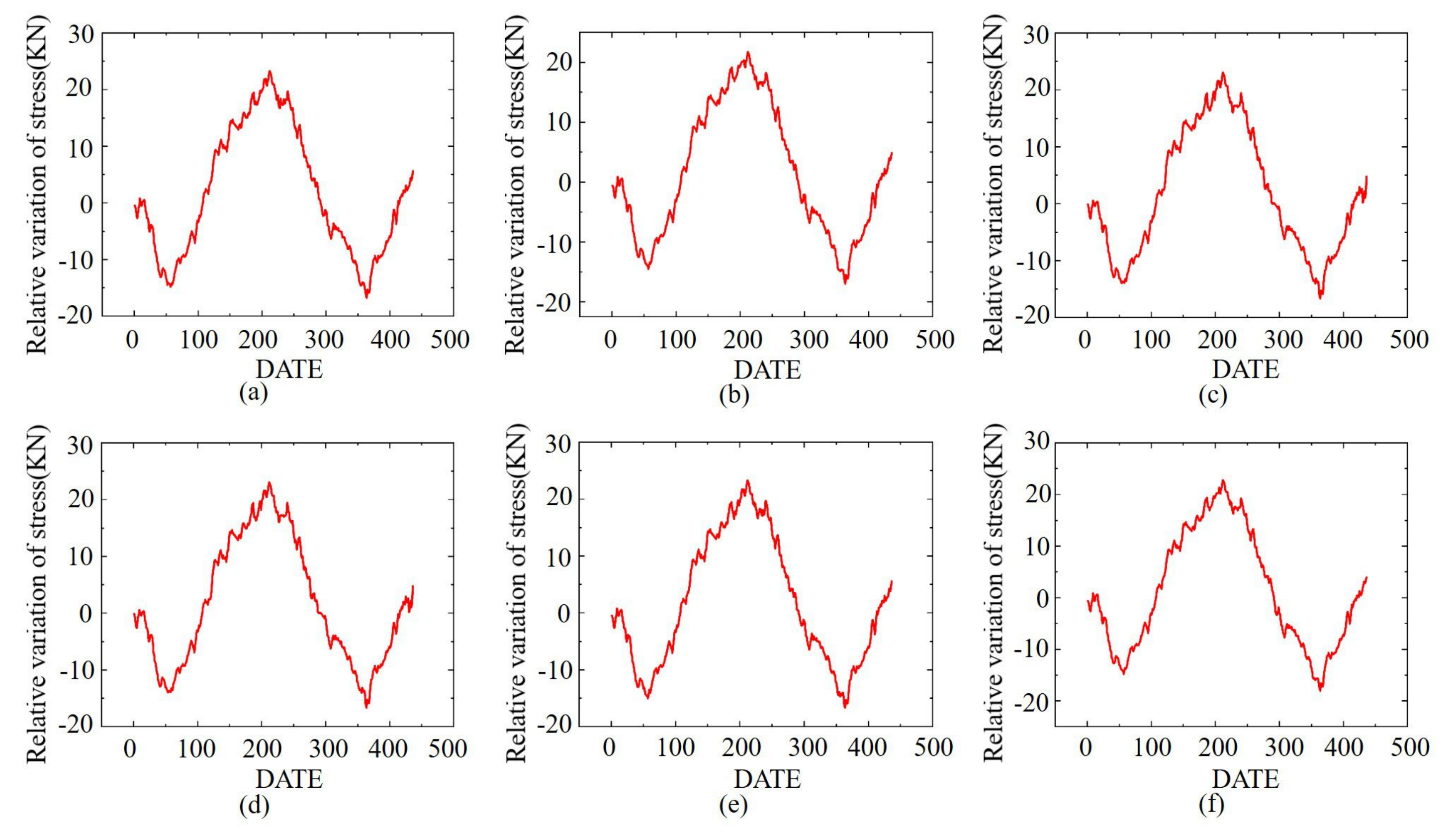}
\caption{The historical performance of six dangerous positions in S2: (a)arch bottom, (b) hance in left-up semicircle, (c) hance in left-low semicircle, (d) hance in right-low semicircle, (e) arch crown,  and (f) hance in right-up semicircle.}
\label{fig:S2_appli}
\end{figure}

\clearpage

\begin{figure}[H]
\centering\includegraphics[width=14cm]{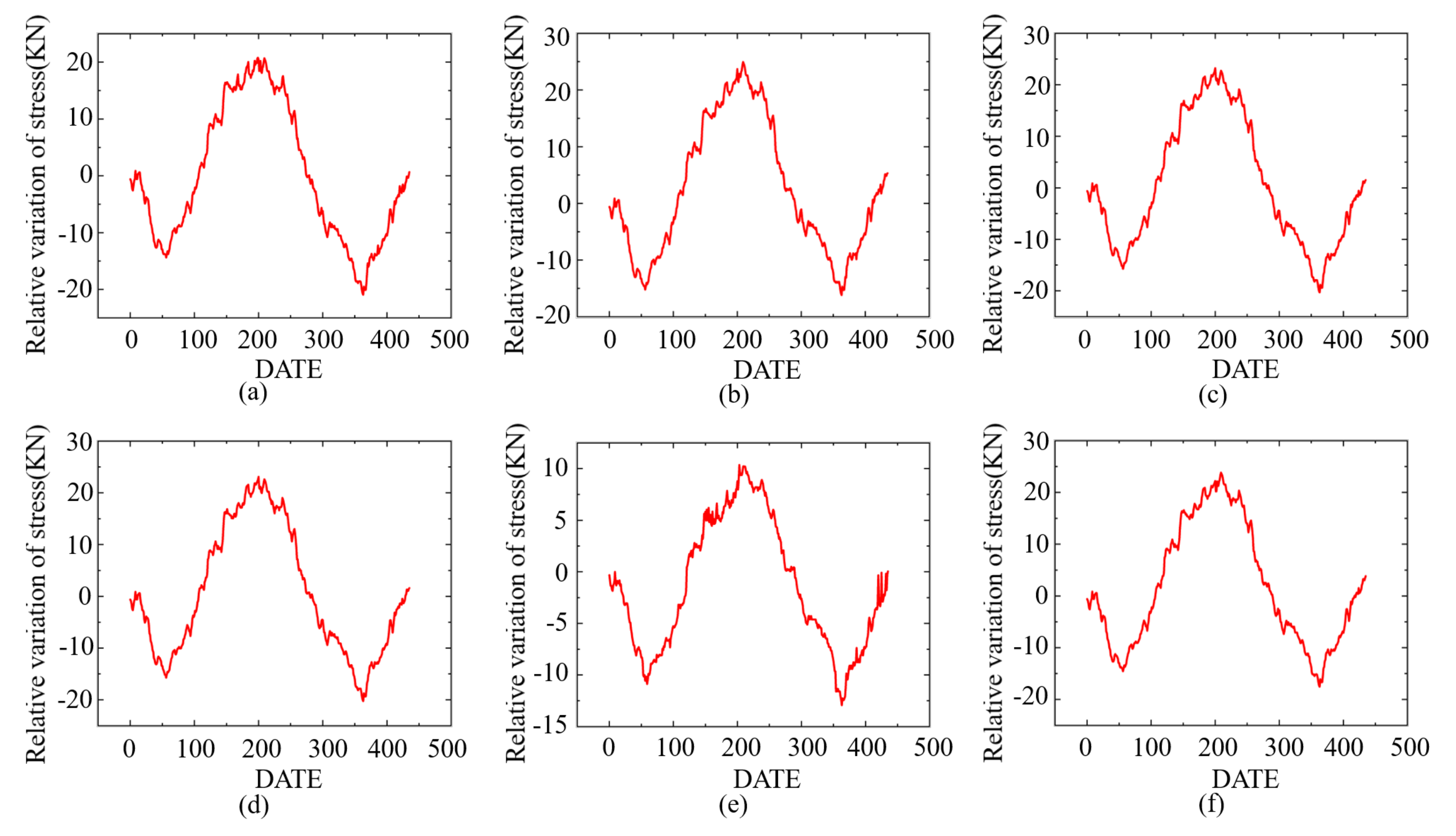}
\caption{The historical performance of six dangerous positions in S4: (a)hance, (b) hance in right-up semicircle, (c) arch bottom, (d) arch bottom, (e)hance in left-low semicircle,  and (f) hance in left-up semicircle.}
\label{fig:S4_appli}
\end{figure}

\clearpage

\begin{figure}[H]
\centering\includegraphics[width=14cm]{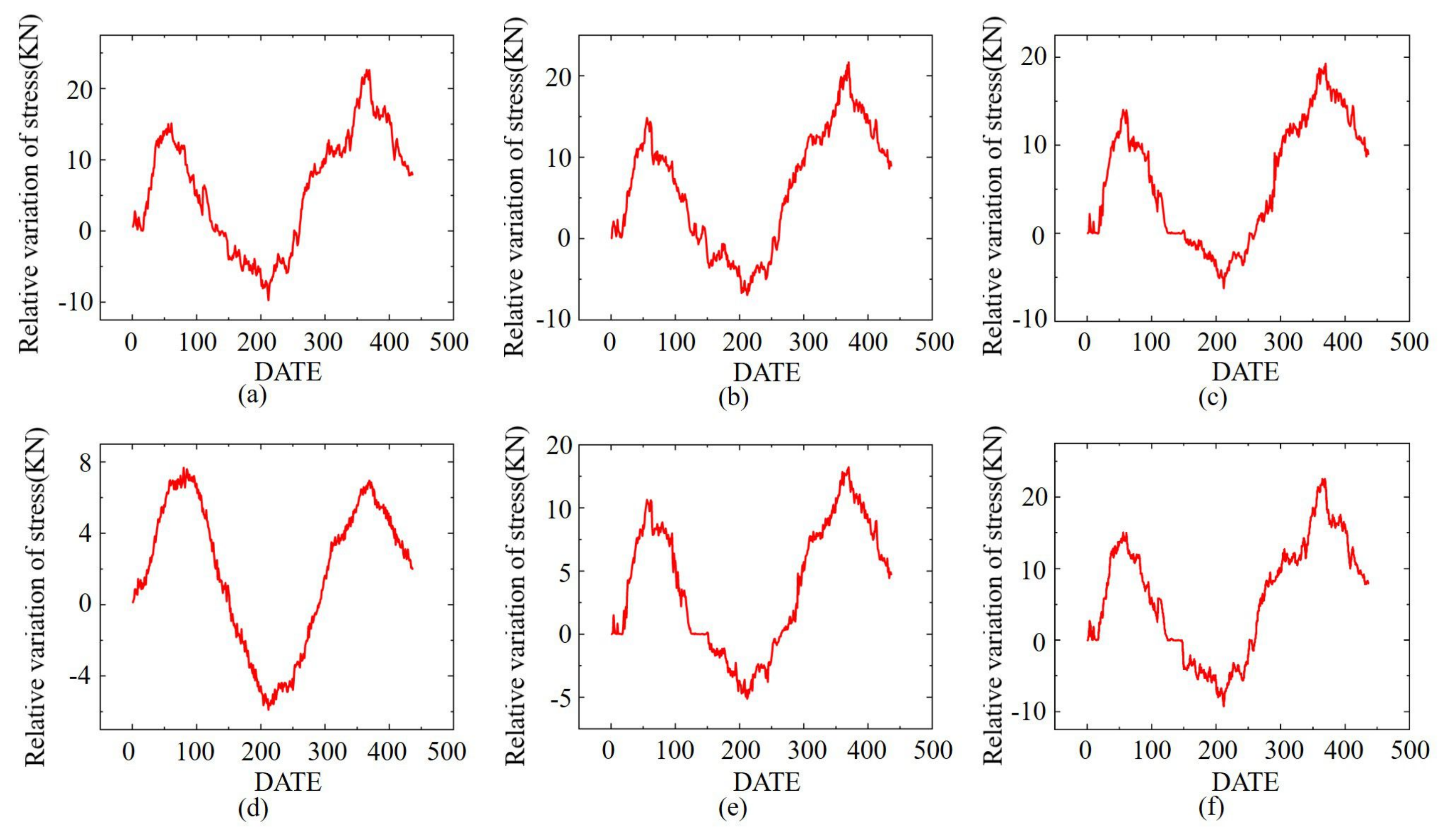}
\caption{The historical performance of six dangerous positions in S9: (a)hance in right-up semicircle, (b) hance, (c)arch bottom, (d) hance in left-low semicircle, (e) arch bottom,  and (f) arch crown.}
\label{fig:S9_appli}
\end{figure}

%% file: Tables.tex
\section *{Appendix: Tables}
\begin{table}[H]
    \centering
    \caption{SGD for NMF}
    \begin{tabular}{l}
    \hline
    \textbf{Algorithm: SGD for NMF}\\ \textbf{Input: the original matrix $X$, iteration steps P, learning rate $\alpha$} \\ \textbf{Output: Two matrices $U$ and $V$}
    \\
    \hline
    randomly init $U$;\\
    randomly init $V$;\\
  step=0;\\
  while step $<$ P:\\
\hspace{1em}for m in range($X.rows$):\\
  \hspace{1em}\hspace{1em}for n in range($X.columns$):\\
    \hspace{1em}\hspace{1em}\hspace{1em}if $X$[m,n] is non-empty:\\
      \hspace{1em}\hspace{1em}\hspace{1em}\hspace{1em}caculate error E[m,n]=$X$[m,n]-dot($U$[m,:],$V$[:,n].T);
      \\
      \hspace{1em}\hspace{1em}\hspace{1em}\hspace{1em}for h in range(H):\\
        \hspace{1em}\hspace{1em}\hspace{1em}\hspace{1em}\hspace{1em}$U$[m,h]=max(0,$U$[m,h]+ $\alpha$*2*E[m,n]*$V$[h,n]);\\
        \hspace{1em}\hspace{1em}\hspace{1em}\hspace{1em}\hspace{1em}$V$[h,n]=max(0,$V$[h,n]+ $\alpha$*2*E[m,n]*$U$[m,h]);\\
    \hspace{1em}step++;\\
    return $U$,$V$;\\
    \hline
    \end{tabular}

    \label{sgd}
\end{table}
% \begin{table}[H]
%     \centering
%     \caption{SGD for NMF}
%     \begin{tabular}{l}
%     \hline
%     \textbf{\textcolor{blue}{Algorithm: SGD for NMF}}\\ \textbf{Input: the original matrix $X$, iteration steps P, learning rate $\alpha$} \\ \textbf{Output: Two matrices $U$ and $V$}
%     \\
%     \hline
%     randomly init $U$;\\
%     randomly init $V$;\\
%   step=0;\\
%   while step $<$ P:\\
% \hspace{1em}for m in range($X.rows$):\\
%   \hspace{1em}\hspace{1em}for n in range($X.columns$):\\
%     \hspace{1em}\hspace{1em}\hspace{1em}if $X$[m,n] is non-empty:\\
%       \hspace{1em}\hspace{1em}\hspace{1em}\hspace{1em}caculate error E[m,n]=$X$[m,n]-dot($U$[m,:],$V$[:,n].T);\\
%       \hspace{1em}\hspace{1em}\hspace{1em}\hspace{1em}for h in range(H):\\
%         \hspace{1em}\hspace{1em}\hspace{1em}\hspace{1em}\hspace{1em}$U$[m,h]=max(0,$U$[m,h]+ $\alpha$*2*E[m,n]*$V$[h,n]);\\
%         \hspace{1em}\hspace{1em}\hspace{1em}\hspace{1em}\hspace{1em}$V$[h,n]=max(0,$V$[h,n]+ $\alpha$*2*E[m,n]*$U$[m,h]);\\
%     \hspace{1em}\textcolor{blue}{step++;}\\
%     \textcolor{blue}{return $U$,$V$;}\\
%     \hline
%     \end{tabular}

%     \label{sgd}
% \end{table}

\clearpage

\begin{table}[H]
    \centering
        \caption{Mechanical parameters of surrounding rock}
    \begin{tabular}{l c c}
    \hline
    \textbf{Ground types} & \textbf{Lateral pressure coefficient} & \textbf{Unite weight}
    \\
    \hline
    Silt & 0.43 & 19.4\\
    Fine sand & 0.40 & 19.3\\
    Silt clay & 0.65 & 18.6\\
    Gravel & 0.25 & 20.6\\
    Gravel sand & 0.30 & 20.3\\
    Weather siltstone & 0.14 & 19.2\\
    \hline
    \end{tabular}

    \label{tab:s4}
\end{table}

\clearpage

\begin{table}[htbp]
    \centering
 
    \caption{Performance on three metrics}
    \linespread{4}
    % \fontsize{7}{8}\selectfont 
    \setlength{\tabcolsep}{6.0pt}{
      \begin{tabular}{c|c c c|c c c}
      \hline
      \multirowcell{2}{Test points} & \multicolumn{3}{c|}{Deduce with bc}& \multicolumn{3}{c}{Deduce without bc}  \\
  \cline{2-7}          & RMSE    & MAE  & PCC & RMSE    & MAE     & PCC \\
      \hline
     S2-B2-inner & 0.4247 & 0.2814 & 0.9995                &1.3587	&0.8249	 &0.9919
                \\
     S2-F-middle & 0.1961 & 0.1284 & 0.9998	
                 &0.8496	&0.6275	&0.9980
                \\
     S2-L1-outer & 0.9320 &  0.7208 & 0.9975	            &1.0015	&0.7836	&0.9951               \\
     S4-B1-inner & 0.6490 & 0.5040
             &0.9989  &0.9732	&0.8207	&0.9979
              \\
     S4-F-middle & 0.8982 & 0.8153 & 0.9950	               &2.1895	&1.7751	&0.9883
              \\
      S4-B3-outer & 0.7061 & 0.5831 & 0.9835	&1.0074	&0.8291		&0.9771\\
      S9-F-inner & 2.0669 & 1.6252 & 0.9684 &6.8944 &4.5032 &0.6781 \\
S9-B7-middle & 0.8650 & 0.6466 & 0.9957 &1.8149 &1.2830 &0.9695\\
S9-L1-outer & 1.9246  & 1.6331 & 0.9433 &5.4977 &4.7329 &0.0028\\

      \hline
      \end{tabular}}
    \label{tab:res}
 \end{table}
 
\clearpage